\documentclass{article}
\usepackage[utf8]{inputenc}
\usepackage{authblk}
\usepackage{setspace}
\usepackage[margin=0.85in]{geometry}
\usepackage{graphicx}
\graphicspath{ {./figures/} }
\usepackage{subcaption}
\usepackage{lineno}
\usepackage{amsmath, amssymb, amsfonts}
\usepackage{array}
\usepackage{url}
\usepackage{xurl}
\usepackage{booktabs}
\usepackage{multirow}
\usepackage{algorithm}
\usepackage{algpseudocode}
\usepackage[table]{xcolor}
\usepackage{textcomp}
\usepackage{comment}
\definecolor{darkblue}{RGB}{0,0,139}
\newcommand{\deltacol}[1]{\textcolor{darkblue}{\textbf{#1}}}
\usepackage{caption}
\newcolumntype{P}[1]{>{\centering\arraybackslash}p{#1}}
\usepackage[style=nejm,
citestyle=numeric-comp,
sorting=none]{biblatex}
 \title{Can Synthetic Data Overcome the Generalization Limits of AI-Based Flower and Pod Detection Across Cowpea Breeding Genotypes and Environments?}
\author[1*$\dag$]{Hamid Kamangir}
\author[2]{Jonathan Berlingeri}
\author[1]{Earl Ranario}
\author[2]{Isaac Kazuo Uyehara}
\author[1]{Lars Lundqvist}
\author[1]{Heesup Yun}
\author[2]{Christine H. Diepenbrock}
\author[2]{Brian N. Bailey}
\author[1]{J.Mason Earles}
\affil[1]{Department of Biological and Agricultural Engineering, University of California Davis, Davis, CA, USA.}
\affil[2]{Department of Plant Science, University of California Davis, Davis, CA, USA.}
\affil[*]{Address correspondence to: hkamangir@ucdavis.edu}
\date{}
\begin{document}
\maketitle
\begin{abstract}
High-throughput phenotyping requires AI-enabled computer vision models that generalize across genotypes, locations, and growing seasons, yet such models often lose accuracy under new conditions. Annotating real imagery for every genotype-by-environment (G x E) combination a breeding program encounters is prohibitively expensive. We quantify how G x E shifts affect AI-based detection of cowpea flowers and pods across two California locations and two growing seasons. Flower detection mAP@50 fell from 76.3\% to as low as 50.6\% under unseen shifts, and pod detection was more sensitive. Feature-space and image-quality diagnostics confirmed these losses track measurable distributional shifts. Because closing this gap with real data alone is not practical, we test whether synthetic imagery, rendered from a procedural 3D cowpea model, can substitute for that annotation burden. Synthetic supervision alone improved over pretraining but remained limited by a domain gap driven by camera image formation, not scene content. A domain-gap-aware camera-realism augmentation strategy, optimized against measured real-image statistics via Wasserstein distance, narrowed this gap, and a linear HDR representation converted a smaller measured gap into a larger detection gain than an 8-bit representation. Optimized HDR synthetic data combined with as few as five real images matched or exceeded the real-data baseline for spatial generalization, and pod detection benefited most at the lowest shot counts, with more modest gains under temporal shift. These results show that synthetic data can overcome the generalization limits of AI-based flower and pod detection, but only when the domain gap is measured and optimized rather than assumed away.
\end{abstract}
\section{Introduction}
High-throughput plant phenotyping (HTP) has become a critical component of modern crop improvement by enabling rapid and quantitative characterization of plant traits at scales required for contemporary breeding programs \cite{nguyen2025advancing}. Accurate phenotypic measurements are essential for identifying superior genotypes, understanding trait--environment interactions, and accelerating genetic gain. In legume crops such as cowpea (\textit{Vigna unguiculata [L.] Walp}), reproductive traits including flower emergence and pod development are closely associated with yield formation, stress response, and adaptation to diverse environmental conditions \cite{oo2022morphological,mohammed2025quantitative}. Consequently, accurate and timely measurement of these traits is important for genotype evaluation, breeding decisions, and crop management. However, conventional phenotyping approaches remain labor-intensive, subjective, costly, and difficult to scale across large breeding populations, multiple locations, and successive growing seasons \cite{pratap2019using, reynolds2020breeder, gao2025enhancing}.

The emergence of HTP systems has provided a pathway to overcome these limitations by combining imaging technologies, sensing platforms, and automated data analysis to measure plant traits at unprecedented scales. Recent advances in artificial intelligence (AI) and computer vision have further accelerated this transition, enabling automated extraction of phenotypic information directly from field imagery \cite{wang2025artificial, chang2026ai, yang2026three, wang2026artificial}. Deep learning models, particularly convolutional neural networks and object detection frameworks, have demonstrated strong performance across a wide range of agricultural applications, including disease detection, yield estimation, organ segmentation, and reproductive trait identification \cite{mohanty2016using, wang2021real, chang2026ai, yang2026three}. Among these approaches, YOLO-based architectures have gained widespread adoption because of their high detection accuracy, computational efficiency, and suitability for real-time deployment \cite{khanam2024yolov11, choudhary2024novel, ayalde2024ai, boddepalli2025ai}. These developments have significantly reduced the burden of manual annotation and accelerated data processing, creating new opportunities for scalable phenotyping pipelines that support breeding programs \cite{lasdun2024participatory,wojcik2024harnessing}.

Applications of deep learning to reproductive trait detection have shown promising results for identifying soybean flowers \cite{zhu2022exploring}, pods, and related yield components under field conditions \cite{dias2018apple, riera2021deep, yang2025deep}. Nevertheless, a major challenge remains: most AI-based phenotyping systems are developed and evaluated within relatively narrow experimental settings and often assume that training and deployment data originate from similar environments \cite{dias2018apple, riera2021deep, zhu2022exploring, choudhary2024novel, oppliger2025investigating, boddepalli2025ai}. Agricultural systems are inherently heterogeneous, with variability arising from both genotype and environment across spatial and temporal scales. Differences in genetic background, location, growing season, management practices, canopy structure, illumination, developmental stage, and image quality introduce substantial variability into field imagery and create distribution shifts between training and deployment domains \cite{oo2022morphological,hartley2021domain}. As a result, models that perform well within a single experiment frequently experience substantial performance degradation when applied to new environments or breeding populations.

From a breeding perspective, this limitation represents a significant barrier to adoption. Breeding programs routinely conduct multi-environment trials and require phenotyping tools that operate reliably across locations, years, and genetic backgrounds while remaining cost-effective to deploy \cite{araus2014field, cobb2019enhancing}. Models that require extensive retraining, and correspondingly extensive re-annotation, for every new trial are difficult to scale and limit the long-term sustainability of AI-driven phenotyping systems. Synthetic imagery, generated on demand from a procedural or physically based plant model rather than collected and hand-labeled in the field, offers a potentially attractive path toward this scalability by decoupling training-data volume from real-world data-collection cost, provided the resulting synthetic-to-real domain gap can be reliably closed. Consequently, improving generalization across genotype-by-environment ($G \times E$) variation, and determining whether synthetic supervision can substitute for the real-world diversity that generalization otherwise requires, has become one of the central challenges in agricultural computer vision.

The effectiveness of synthetic supervision, however, is constrained by a well-documented sim-to-real domain gap. This gap is largely an image-formation gap rather than a content gap. Even a geometrically faithful, physically based renderer produces an idealized image. It has uniform exposure, no sensor noise, and no compression artifacts, features that a real camera's sensor and image-signal-processing (ISP) pipeline never produces on its own \cite{tobin2017domain, tremblay2018training}. Two broad strategies exist for closing this gap. The first is to increase rendering photorealism until simulated output is statistically indistinguishable from a real photograph. This approach is expensive and requires continual re-rendering as target cameras or field conditions change. In practice, this strategy never fully succeeds because a renderer without an explicit sensor-noise and ISP model will not spontaneously reproduce a specific camera's compression artifacts or white-balance error. The second, and the strategy this study builds on, is \textit{domain randomization}: rather than pursuing photorealism, a broad, controllable family of post-render transforms is applied so that the training distribution spans the statistical space real cameras actually produce. Tobin et al. \cite{tobin2017domain} established this principle for sim-to-real transfer in robotics, and Tremblay et al. \cite{tremblay2018training} extended it specifically to object detection, showing that non-photorealistic domain randomization, combined with a modest amount of real fine-tuning data, outperformed training on photorealistic synthetic data alone. An alternative, image-to-image domain-transfer approach (e.g., CycleGAN-style unpaired translation \cite{zhu2017unpaired}) can also narrow this gap, but introduces a second large model to train and validate and carries a known risk of hallucinating or shifting content near object boundaries, a risk that directly corrupts detection labels in a way that geometry-preserving, parametric augmentation does not.

Not every augmentation strategy is equally suited to this problem, however. The augmentation built into standard object-detection training pipelines (generic color jitter, mosaic tiling, and similar transforms) is a domain-agnostic regularizer. It is designed to reduce overfitting to a single training distribution, not to close a \textit{measured} gap between two specific, named image domains. Its fixed, hand-picked ranges carry no information about how a given synthetic renderer's output statistically differs from a given real camera's output. Closing a measured domain gap instead requires augmentation parameters that are fit to data, in the spirit of data-driven augmentation-policy search \cite{cubuk2019autoaugment}, using an objective that meaningfully captures distributional similarity. A further consideration specific to synthetic rendering is bit depth. Standard 8-bit synthetic renders are already the output of a baked-in, irreversible tone-mapping and exposure decision. Once quantized to 8 bits, that decision cannot be revisited. Retaining the renderer's native linear float32 high-dynamic-range (HDR) output, commonly stored as EXR, instead preserves the radiometric information needed to make a physically accurate, data-driven exposure and white-balance correction after the fact \cite{debevec1997recovering}. This follows the same spirit as prior work showing that operating in a raw, physically meaningful image representation, rather than an already-processed, nonlinear one, gives a model access to information a camera's own ISP has otherwise discarded \cite{brooks2019unprocessing}.

In this work, we explicitly frame phenotyping generalization through the lens of genotype-by-environment interactions. Using cowpea flower and pod detection as a case study, we design controlled experiments that isolate genotype effects, environmental effects, and their interactions through combinations of seen and unseen genotypes and environments. In addition, we analyze distribution shifts using both representation-level and image-quality metrics to better understand the mechanisms underlying model failure under domain shift. Together, these experiments provide a systematic framework for quantifying generalization, identifying the dominant drivers of performance degradation, and establishing the empirical case for the central question this study asks: whether, and under what conditions, synthetic data can overcome the resulting generalization limits.

Building directly on that motivation, our central contribution is not synthetic image generation on its own. It is a domain-gap-aware augmentation-optimization framework that determines how effectively synthetic imagery transfers to real conditions. Rather than hand-selecting augmentation parameters or relying on a detector's built-in augmentation, we search for augmentation configurations that explicitly minimize the distributional distance between augmented synthetic and real imagery. This distance is measured jointly through image-quality statistics and representation-level (embedding) similarity. We adopt the Wasserstein (Earth Mover's) distance \cite{rubner2000earth} as the core distributional objective, following the same underlying rationale that motivates its use as a training signal in Wasserstein GAN \cite{arjovsky2017wasserstein}. It captures the full shape of a distributional difference rather than only its mean and variance, which matters because image-quality statistics are not well described by their first two moments alone. This optimization is performed entirely without target-domain object annotations, relying only on image statistics and unlabeled real imagery. It is conceptually related to, though computationally cheaper than, reinforcement-learning-based simulation-parameter optimization against downstream task accuracy \cite{ruiz2019learning}. We evaluate this framework across both 8-bit and HDR (EXR) synthetic representations, and quantify how the resulting optimized synthetic supervision interacts with few-shot adaptation. This lets us assess the extent to which synthetic data, when properly aligned to the real domain, can replace or complement real annotation under unseen genotype and environmental conditions.

These objectives motivate the following hypotheses, which guide the experimental design and analysis throughout this study.
\subsection{Hypotheses}
To evaluate the generalizability of deep learning-based flower and pod detection under genotype and environmental variation, we define environment ($E$) as the combination of location and year. Based on agronomic knowledge and expected visual variability across growing conditions, we organize our hypotheses into three categories: genotype-by-environment (G$\times$E) distribution shift, synthetic-to-real transfer, and trait-specific sensitivity.
\paragraph{Genotype-by-Environment (G$\times$E) Distribution Shift}
\begin{itemize}
    \item \textbf{H1 (Genotype Dependency):} Detection performance degrades when models are evaluated on unseen genotypes due to genotype-specific morphological and phenological differences.
    \item \textbf{H2 (Environmental Domain Shift):} Changes in environment (location and/or year) introduce domain shifts that negatively impact detection performance, with compounded effects when both genotype and environment are unseen.
\end{itemize}
\paragraph{Synthetic-to-Real Transfer}
\begin{itemize}
    \item \textbf{H3 (Synthetic-Real Complementarity):} Incorporating synthetic training data and/or few-shot fine-tuning using a small number of labeled real samples from a target genotype-environment domain improves performance under domain shift. Combining synthetic supervision with limited real supervision outperforms either source alone, substantially reducing the number of real annotations required to reach a given performance level.
    \item \textbf{H4 (Augmentation Fidelity Governs Synthetic Transfer):} The extent to which synthetic supervision transfers to real conditions is governed primarily by how closely the augmented synthetic image distribution matches the real image distribution, rather than by the presence of synthetic data alone. Specifically, we hypothesize that (i) camera-realism augmentation whose parameters are explicitly optimized against measured real-image statistics will outperform both unaugmented synthetic data and augmentation using hand-set, non-optimized parameters, and (ii) rendering and augmenting synthetic imagery in linear float32 HDR (EXR), which preserves radiometric information that 8-bit rendering has already discarded, will yield a larger and more reliable improvement than augmenting 8-bit imagery alone.
\end{itemize}
\paragraph{Trait-Specific Sensitivity}
\begin{itemize}
    \item \textbf{H5 (Object-Specific Sensitivity):} Pod detection is more sensitive to genotype and environmental shifts than flower detection because of greater structural variability, occlusion, image-quality sensitivity, and growth-stage dependence.
\end{itemize}

\section{Materials and Methods}
\subsection{Dataset Overview}
\subsubsection{Real Dataset}
\begin{figure*}[ht!]
    \centering
    \includegraphics[width=0.98\textwidth]{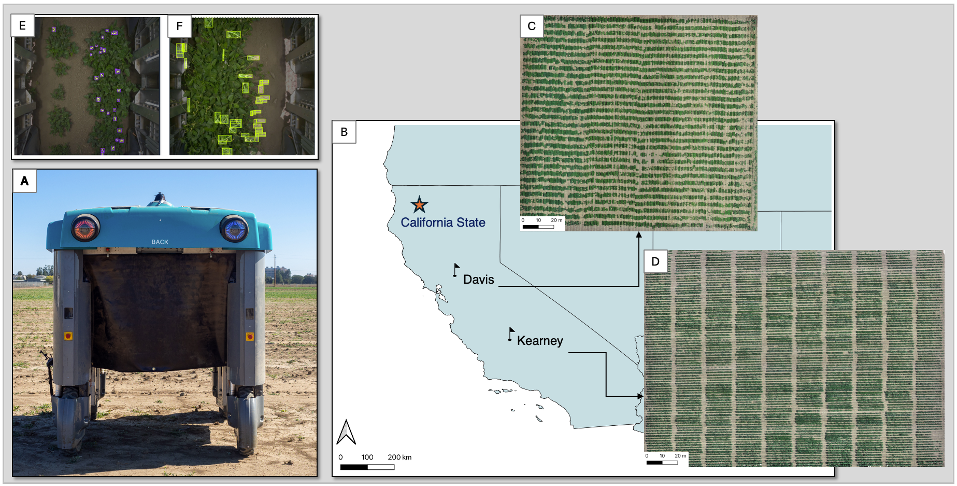}
    \caption{Overview of the data collection system, study locations, and detection outputs. (A) T4 rover platform used for field image acquisition. (B) Map of California showing the Davis and Kearney experimental sites. (C) Aerial view of the Davis experimental field. (D) Aerial view of the Kearney experimental field. (E) Example flower detection output from T4 imagery. (F) Example pod detection output from T4 imagery. Note that the aerial images depict the complete experimental fields at each site, including both the MAGIC population used in this study and additional breeding populations (e.g., interspecific populations) that were not included in the analyses.}
    \label{fig:area}
\end{figure*}

The dataset used in this study was collected from multi-environment cowpea field experiments conducted at two research locations, Davis, California (Figure~\ref{fig:area}.C) and Kearney Agricultural Research Station in Parlier, California (Figure~\ref{fig:area}.D), during the 2022 and 2023 growing seasons (Figure~\ref{fig:area}). Note that the aerial images depict the complete experimental fields at each site, including both the MAGIC population used in this study and additional breeding populations (e.g., interspecific populations) that were not included in the analyses. Images were acquired using the Google Mineral T4 rover shown in Figure~\ref{fig:area}A, a manually-driven platform equipped with an Intel i7-based Linux computer, network-RTK GPS, and six $2592\times2048$ RGB cameras plus two NIR cameras mounted 1.5 m above ground. The rover collected data at a 2 m/s driving speed and captured images at 10 Hz. Figure~\ref{fig:area} E and F show two sample outputs of flower and pod detection, respectively, from T4 imagery.

\paragraph{Real Image Sampling and Labeling}
To ensure comparable developmental stages across environments, image selection was restricted to the June–August period, corresponding to the primary flowering and early pod development stages of cowpea. Images were sampled across the available multi-parent advanced generation inter-cross (MAGIC) population \cite{huynh2018multi} grown at the Davis and Kearney field sites. The MAGIC population originally consists of 305 lines. Only 168 MAGIC genotypes, along with selected parental lines, were evaluated in these experiments because the remaining lines were expected to flower too late to reach maturity under the field conditions. Within this subset, image selection focused on capturing visible flower and pod objects across genotypes and environments. As a result, not all genotypes were represented equally in the final detection datasets, since reproductive expression varied across genotypes, locations, and years. This sampling strategy preserved natural phenological variability while ensuring that the dataset contained sufficient flower and pod instances for robust model training and evaluation across genotype–environment domains.

\begin{figure*}[ht!]
    \centering
    \includegraphics[width=0.98\textwidth]{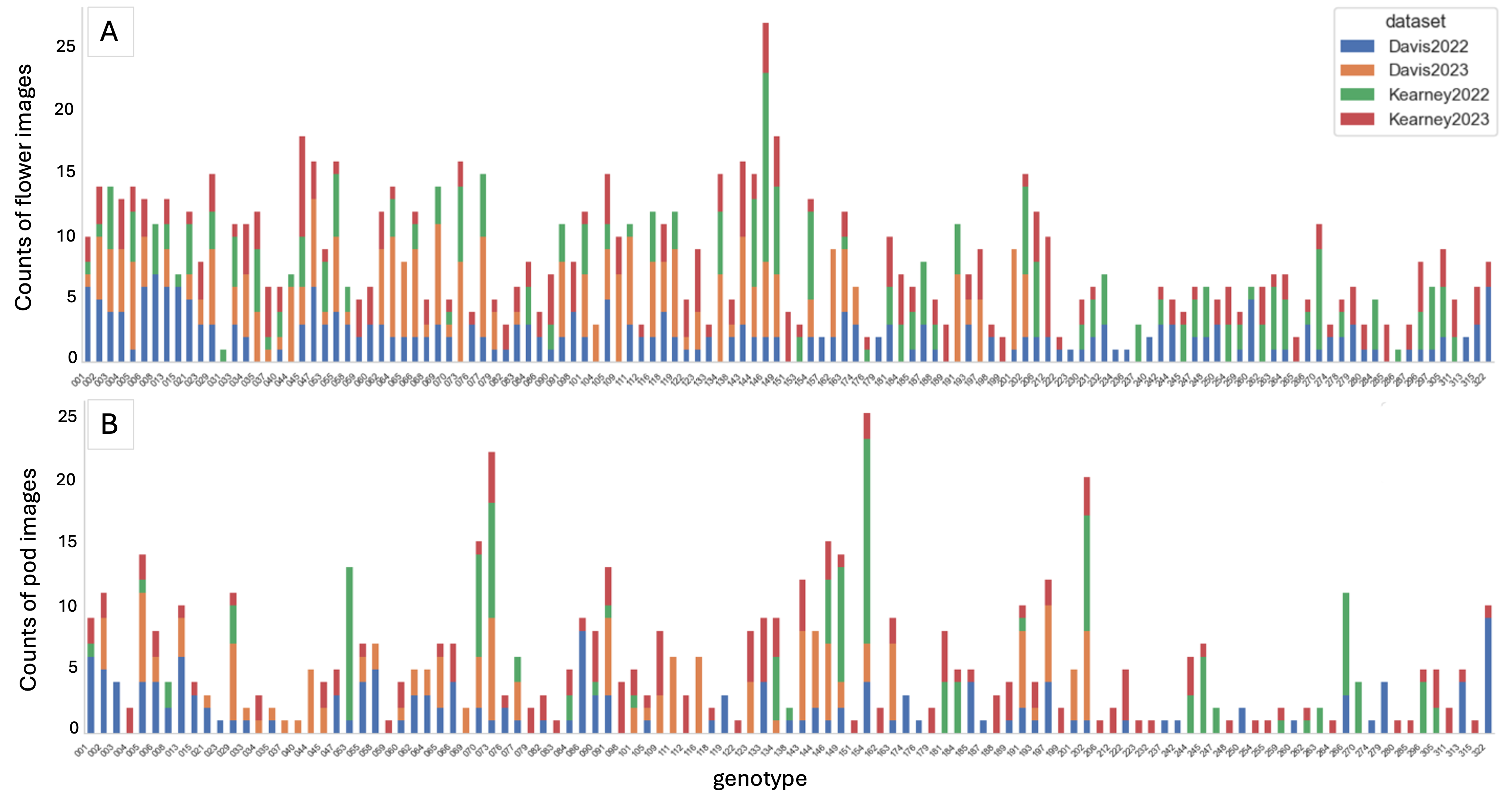}
    \caption{    Distribution of flower (A) and pod (B) detection images across cowpea genotypes for four environment--year datasets.
    Each bar represents the number of images collected for a given MAGIC genotype, colored by dataset.}
    \label{fig:flower_pod_count}
\end{figure*}

The training images were cropped to a spatial resolution of $640 \times 640$ pixels and annotated by domain experts in plant breeding using the Roboflow platform \cite{roboflow}. All pods in the image set were labeled except very immature pods (<~2 inches in length), as these could not be reliably distinguished developmentally from mature flowers due to variability in flower abscission. All flowers were labeled except green buds and highly senescent flowers that had begun to decompose (brown in color), as these were difficult to distinguish from pods. The final dataset contains images and annotations from four environment–year combinations: Davis 2022, Davis 2023, Kearney 2022, and Kearney 2023. For flower detection, the dataset consists of 992 images and 6,926 annotated flower instances, distributed across 101 genotypes in Davis 2022, 68 genotypes in Davis 2023, 72 genotypes in Kearney 2022, and 92 genotypes in Kearney 2023. For pod detection, the dataset contains 596 images and 4,998 annotated pod instances, covering 48, 52, 55, and 60 genotypes for Davis 2022, Davis 2023, Kearney 2022, and Kearney 2023, respectively.

Figures \ref{fig:flower_pod_count} A and B further illustrate the distribution of images across genotypes for the flower and pod datasets, respectively. The bar charts show the number of images associated with each genotype across the four environment–year combinations. While the sampling procedure aimed to maintain balanced representation across genotypes, natural variations in flowering and pod development, as well as differences in field conditions and phenological timing across environments, resulted in uneven distributions. This variability is intentionally preserved, as it reflects realistic agricultural scenarios where genotype expression and plant architecture differ across environments. Consequently, the dataset provides a suitable benchmark for evaluating how genotype and environmental variation influence deep learning model generalization in field-based plant detection tasks.
\subsubsection{Synthetic Dataset}

Recent studies have highlighted the potential of synthetic data as a scalable alternative to manual annotation, offering virtually unlimited labeled samples for training while reducing both labeling costs and human-induced errors \cite{xu2023handsoff}. Building on this motivation, we explore the role of synthetic imagery in enhancing model generalization under genotype–environment distribution shifts. Specifically, we design a set of experiments that combine synthetic data with few-shot learning to evaluate how effectively synthetic supervision can transfer to real-world conditions.
To support this investigation, we generated synthetic datasets for each detection task (flower and pod), with 1,000 images per task, using Helios \cite{bailey2019helios, lei2024simulation}. Helios is an open-source 3D biophysical modeling framework for simulating plant structures and their surrounding environments (available at \url{https://github.com/PlantSimulationLab/Helios}). It leverages a physics-based radiation modeling approach to produce both realistic RGB imagery and corresponding ground-truth annotations, including segmentation and object detection masks. Because Helios' radiation model computes native linear radiance rather than an already tone-mapped image, we retain the renderer's output in two parallel representations for every scene: a standard 8-bit RGB rendering, and the underlying linear float32 high-dynamic-range (HDR) radiance buffer, stored as EXR. As detailed in Section~\ref{sec:aug}, this HDR representation is what makes a physically motivated, data-driven camera-realism augmentation possible, since it preserves radiometric information an 8-bit rendering has already discarded.

\begin{figure*}[ht!]
    \centering
    \includegraphics[width=0.95\textwidth]{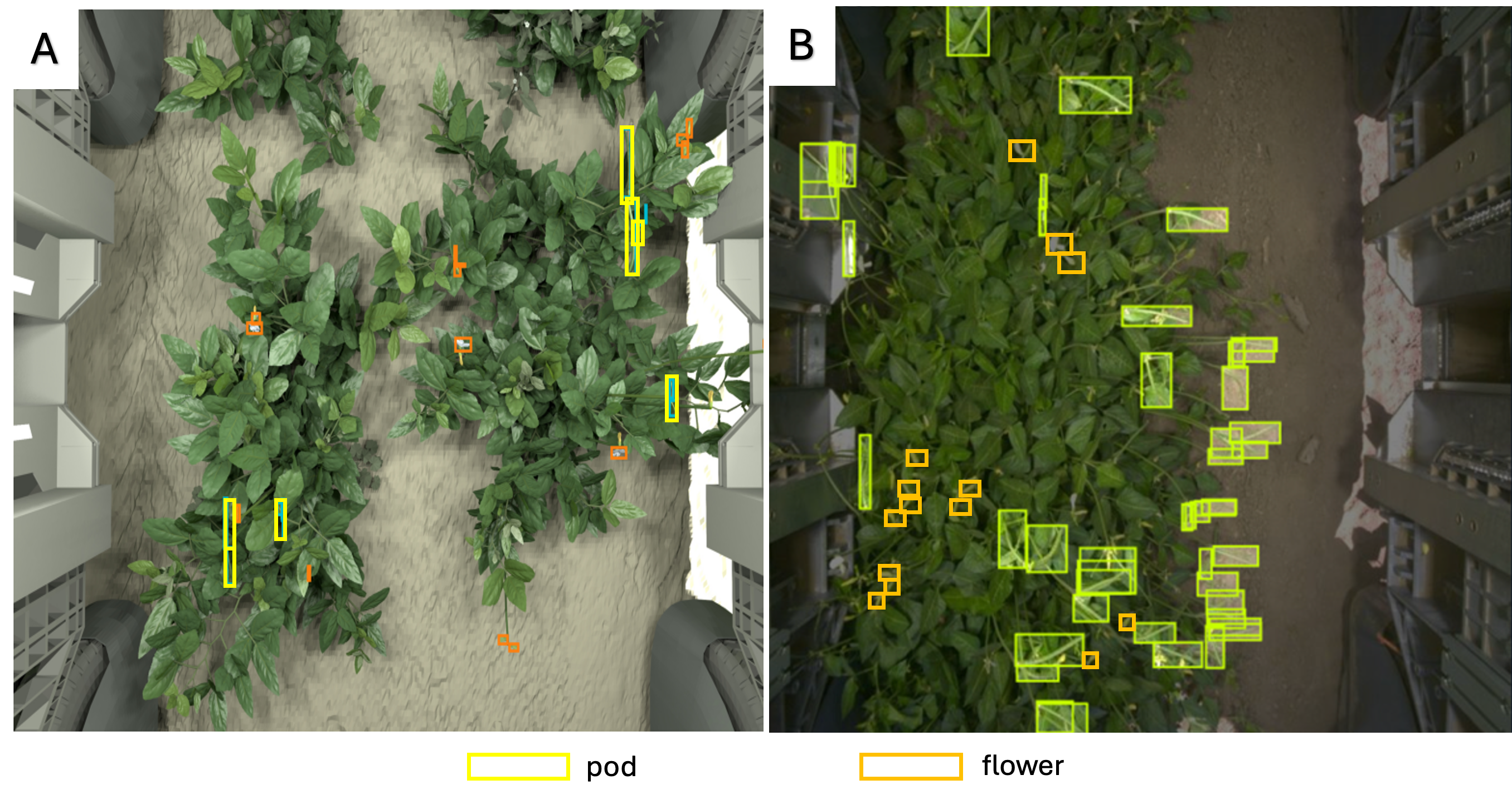}
    \caption{Synthetic, 8-bit (A, left) and Real (B, right) Image Samples.}
    \label{fig:synthetic_sample}
\end{figure*}

Within this framework, cowpea plants were procedurally generated at growth stages aligned with field observations using the plant architecture model plugin, which enables parametric control over plant morphology. The rendering pipeline incorporates detailed physical properties, including spectral reflectance and transmittance of scene elements, illumination characteristics, and camera response functions. These parameters were defined using a combination of Helios' spectral libraries and manufacturer specifications for the imaging system. Object-level annotations were created by assigning unique identifiers to each flower and pod instance, allowing Helios to automatically generate pixel-wise instance masks for every image (see Figure \ref{fig:synthetic_sample} A). Further details on the simulation pipeline are provided in \cite{lei2024simulation}.

\subsection{Modeling and Methodology for Qualitative and Quantitative Analysis}
\begin{figure*}[ht!]
    \centering
    \includegraphics[width=0.95\textwidth]{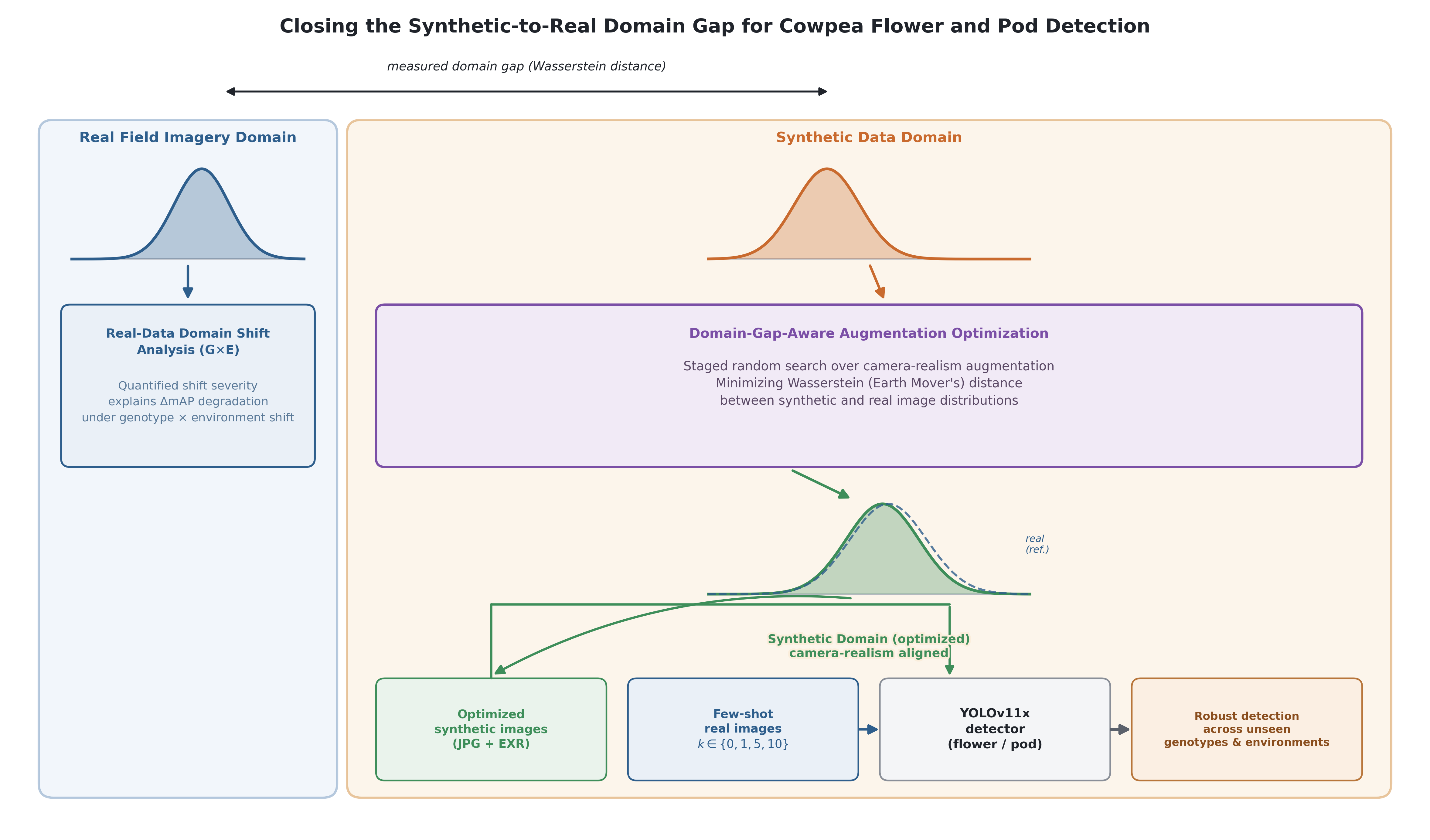}
    \caption{Overview of the study design. Real-data domain shift analysis (left) quantifies genotype-by-environment effects on real imagery, while domain-gap-aware augmentation optimization (right) aligns synthetic renders with real image statistics by minimizing Wasserstein distance. Optimized synthetic images combined with few-shot real supervision train the YOLOv11x detector for robust flower and pod detection across unseen genotypes and environments.}
    \label{fig:domain_gap_schematic}
\end{figure*}

This section and Figure~\ref{fig:domain_gap_schematic} summarize our overall approach. We quantify genotype-by-environment domain shift in real imagery, close the synthetic-to-real gap through optimized camera-realism augmentation, and enable robust few-shot detection across unseen conditions.

\subsubsection{Real Data GxE Training Experiments}
To systematically analyze model behavior, experiments are grouped into four generalization regimes based on whether genotype ($G$) and environment ($E$) are observed during training or held out:
\begin{itemize}
    \item $\mathbf{G_sE_s}$: Seen genotype and seen environment (baseline)
    \item $\mathbf{G_sE_u}$: Seen genotype, unseen environment
    \item $\mathbf{G_uE_s}$: Unseen genotype, seen environment
    \item $\mathbf{G_uE_u}$: Unseen genotype and unseen environment (either location or year)
\end{itemize}
This experimental design enables isolation of genotype effects, environmental effects, and their interaction on detection performance. In these experiments, the seen dataset is used for training and validation, and the unseen dataset is used for test evaluation.

\subsubsection{Methodology for Qualitative Analysis of Domain Shift Effects}
To further understand the mechanisms behind the performance degradation, we analyze the underlying distributional differences between training and test datasets using both representation-level and image-level statistics. Specifically, we quantify shifts in the learned feature space using DINOv2 embeddings and evaluate variations in image quality characteristics that may contribute to domain shifts across genotype–environment conditions. This complementary analysis provides insight into why certain experimental regimes exhibit stronger generalization degradation than others.
Let $X^{tr} = \{x_i^{tr}\}_{i=1}^{N}$ and $X^{te} = \{x_j^{te}\}_{j=1}^{M}$ denote the sets of DINOv2 embeddings extracted from training and test images, respectively. We first measure the embedding centroid shift as the Euclidean distance between the mean embeddings,
\begin{equation}
\Delta_c = \left\lVert \mu_{tr} - \mu_{te} \right\rVert_2,
\end{equation}
where $\mu_{tr} = \frac{1}{N}\sum_{i=1}^{N} x_i^{tr}$ and $\mu_{te} = \frac{1}{M}\sum_{j=1}^{M} x_j^{te}$. This metric captures global displacement of the test distribution relative to the training manifold in representation space. To characterize higher-order distributional differences beyond the centroid, we compute the Wasserstein (Earth Mover's) distance independently for each embedding dimension and report their mean,
\begin{equation}
\Delta_W^{emb} = \frac{1}{D}\sum_{d=1}^{D} W_1\!\left(X^{tr}_{:,d}, X^{te}_{:,d}\right),
\end{equation}
where $W_1(\cdot,\cdot)$ denotes the 1-D Wasserstein distance and $D$ is the embedding dimensionality. This measure captures broader distributional changes including differences in variance, skewness, and multimodality of embedding features. Finally, we quantify downstream generalization loss using relative detection performance degradation compared to the seen–seen baseline,
\begin{equation}
\Delta \mathrm{mAP}@50 =
{\mathrm{mAP}^{base} - \mathrm{mAP}^{exp}},
\end{equation}
where $\mathrm{mAP}^{base}$ and $\mathrm{mAP}^{exp}$ denote baseline and experimental performance, respectively.
In addition to representation-level analysis, we evaluated differences between training and test image distributions using four image statistics: Laplacian sharpness, noise level, brightness mean, and contrast standard deviation. These metrics primarily capture changes in image appearance and acquisition conditions, such as illumination, shadows, soil background, canopy coverage, and sensor response, rather than direct measures of plant physiology. Consequently, they should be interpreted as indicators of visual domain shift that may influence model generalization. These metrics capture variations in illumination conditions, sensor noise, and image clarity that may influence model performance across environments. For each experiment, normalized values of embedding and image quality metrics are visualized using radar plots and their combined polygon areas are used as an aggregate measure of domain shift severity.

\subsubsection{Domain-Gap-Aware Camera-Realism Augmentation}
\label{sec:aug}
A geometrically faithful synthetic renderer does not, by itself, reproduce the statistical signature of a real camera image. A real photograph carries sensor shot and read noise. It also carries an auto-exposure and white-balance decision made by the camera's image-signal processor, lens vignetting, and, for JPEG output, quantization and compression artifacts. A physics-based radiance renderer introduces none of these unless they are explicitly modeled. We treat this as an image-formation gap, distinct from the scene-content gap that improving renderer fidelity addresses. We close it using a domain-randomization-style camera-realism augmentation pipeline \cite{tobin2017domain,tremblay2018training}, applied after rendering rather than during it. Every transform in this pipeline corresponds to a specific, physically motivated camera or ISP-level operation: exposure and brightness scaling, contrast and gamma adjustment, saturation scaling, per-channel white-balance gain, sensor noise injection, blur, unsharp-mask sharpening, lens vignetting, and JPEG-quality simulation. Each transform is a parametric, per-pixel or per-region operation. It is therefore analytically invertible and preserves object geometry exactly, avoiding the label-corruption risk associated with learned image-to-image domain-transfer approaches such as CycleGAN-style translation \cite{zhu2017unpaired}.

We deliberately do not rely on the object detector's built-in augmentation (e.g., mosaic tiling, generic hue/saturation/value jitter, and affine geometric transforms) for this purpose. Such augmentation is a domain-agnostic regularizer. It is designed to improve robustness to variation \textit{within} a single training distribution, not to close a \textit{measured} gap between two specific, named image domains. A fixed, hand-picked color-jitter range carries no information about how our particular synthetic renderer's output statistically differs from our particular real camera domains. We disable this built-in augmentation during synthetic training so that the camera-realism pipeline is the only photometric intervention in effect. This lets us attribute any change in the measured domain gap or in downstream detection performance to that pipeline, rather than confounding it with a second, independently randomized augmentation system.
This augmentation is applied to both the 8-bit and the linear float32 HDR (EXR) synthetic representations described above. For the HDR representation, augmentation is performed directly in linear radiance space. The exposure and white-balance decision that ultimately maps the image to a viewable range is deferred and recomputed per augmented copy, from an automatic highlight-based exposure estimate. This mirrors how a real camera's own auto-exposure algorithm operates at capture time \cite{debevec1997recovering}. It is only possible because the underlying HDR data have not yet been quantized or gamma-encoded. Once an image has been reduced to 8-bit output, this radiometric information is permanently lost, and any post-hoc exposure or white-balance correction is limited to redistributing an already-clipped range \cite{brooks2019unprocessing}. We therefore evaluate both representations explicitly, hypothesizing that the HDR pathway provides additional headroom over 8-bit augmentation alone (H4).

Rather than hand-selecting the parameters of this augmentation pipeline, we treat them as hyperparameters to be fit to data, following the same underlying rationale as data-driven augmentation-policy search \cite{cubuk2019autoaugment}. For each transform family, we perform a staged random search over its parameters, retaining and carrying forward the best configuration found before searching the next family. Each candidate configuration is scored by generating an augmented sample batch and computing the average, per-metric normalized Wasserstein (Earth Mover's) distance \cite{rubner2000earth} between the augmented synthetic and real image distributions. This score combines low-level image-quality statistics (brightness, contrast, sharpness, saturation, and noise characteristics) with high-level embedding-space distances between synthetic and real images. We adopt the Wasserstein distance, rather than a simpler mean and standard-deviation comparison, because it captures the full shape of a distributional difference, including tail behavior that a two-moment comparison can miss. This follows the same rationale that motivates its use as a training signal in Wasserstein GAN \cite{arjovsky2017wasserstein}. Scores are combined across the two available real domains using a worst-domain-aware weighting, so that fit to one real domain cannot be improved at the expense of silently sacrificing the other. From this search, we select the configurations achieving the lowest combined discrepancy score and apply them to generate the final augmented synthetic training sets used in the experiments below. This approach is conceptually related to reinforcement-learning-based optimization of simulation parameters directly against downstream task accuracy \cite{ruiz2019learning}, though substantially cheaper. It optimizes a distributional proxy for domain similarity rather than detection accuracy itself, which we treat as a necessary but not sufficient condition for improved transfer, and evaluate directly against the few-shot detection results reported below.

\subsubsection{Few-Shot Synthetic-Real Training Experiments}
To evaluate how synthetic supervision, augmentation optimization, and limited real annotation interact, we train detection models under a common set of regimes for each target real domain (unseen Kearney and unseen $Y_{2022}$, corresponding to spatial and temporal generalization, respectively). These regimes span three conditions. The first is zero-shot transfer from synthetic data alone, with no real images used during training. The second is synthetic-plus-real training, in which every training batch mixes synthetic images with a fixed, small number of real images ($k \in \{1, 5, 10\}$ shots) drawn from the target domain. The third is real-only few-shot training using the same $k$ real images with no synthetic supervision, serving as a lower-bound reference for how far limited real annotation alone can go. For each shot count and domain, synthetic supervision is further evaluated under three augmentation states: no augmentation (WOA), augmentation using hand-set, non-optimized parameters (WA, internal), and augmentation using the domain-gap-optimized parameters from Section~\ref{sec:aug} (WA, optimized), each evaluated for both the 8-bit (JPG) and HDR (EXR) synthetic representations. Because few-shot performance is sensitive to which specific real images are sampled for a given shot count, each configuration is repeated over 10 independent training episodes with different sampled real images, and we report the mean and standard deviation of detection performance across episodes.

\subsubsection{Model Training Framework for Object Detection}
All experiments were conducted using the YOLOv11 extra-large (YOLOv11x) object detection model initialized from the COCO pre-trained checkpoint (common object context, \cite{lin2014microsoft}). The same model architecture, initialization, and training configuration were used across all experimental scenarios to ensure fair and controlled comparisons. Both flower and pod detection tasks were trained and evaluated within the Roboflow training and evaluation framework, which provided a consistent pipeline for data splitting, augmentation, model optimization, and metric computation including mean average precision (mAP), precision and recall. By fixing the model capacity, pre-training source, and training environment across experiments, observed performance differences can be attributed primarily to variations in genotype and environmental generalization rather than architectural or optimization effects.

\section{Results}
This section quantifies the impact of genotype and environmental distribution shifts on flower and pod detection performance across the proposed generalization regimes (\textit{Quantifying the Effect of G$\times$E Distribution Shifts on Detection Performance}). We then investigate the underlying causes of performance degradation through representation-level and image-quality shift analyses (\textit{Qualitative Analysis of Representation and Image Quality Shifts}), followed by an evaluation of synthetic data and few-shot adaptation strategies for improving robustness under domain shift (\textit{Synthetic Data and Few-Shot Adaptation}). Across all experiments, environmental shifts generally produced larger performance losses than genotype-only shifts, and pod detection was consistently more sensitive to domain shift than flower detection.
\subsection{Quantifying the Effect of G × E Distribution Shifts on Detection Performance}

Table~\ref{table:experiment results} summarizes detection performance across different genotype by environment generalization regimes. For flower detection, the baseline experiment ($G_sE_s$), where both genotype and environment are observed during training, achieves an mAP@50 of 76.3\% with 80.8\% precision and 71.5\% recall. When evaluating on unseen genotypes while keeping environments observed ($G_uE_s$), performance drops modestly to 74.6\% mAP@50, indicating a limited but measurable genotype dependency. In contrast, temporal shifts between years ($T$) produce a larger degradation, reducing performance to 65.7–64.5\% mAP@50. Even stronger effects are observed under spatial shifts between locations ($S$), where mAP decreases to 51.7–67.3\% with $\Delta$mAP values up to $-24.6 \%$, highlighting the substantial visual domain differences between Davis and Kearney environments. When genotype variation is combined with environmental changes, the performance degradation becomes more pronounced. The spatio-genotype shift ($SG$) results in mAP values of 50.6–60.8\% with $\Delta$mAP reaching $-25.7 \%$, while temporal-genotype shifts ($TG$) produce intermediate degradation (58.6–63.1\% mAP). Within-year comparisons show that models trained on 2023 generalize slightly better to 2022 than vice versa, suggesting stronger variability in the 2023 growing season. Similarly, within-location experiments indicate asymmetric transfer between Davis and Kearney, with models trained on Kearney showing better transfer to Davis than the reverse, consistent with location-specific environmental complexity.

\begin{table*}[ht!]
\footnotesize
\centering
\caption{Detection performance across generalization regimes.
Results are reported using $\mathrm{mAP}@50$, Precision, and Recall for flower and pod detection.}
\arrayrulecolor{black}
\resizebox{\textwidth}{!}{%
\begin{tabular}{l p{4.3cm} p{1.1cm} p{0.9cm} p{1.3cm} p{1.3cm} p{1.3cm} p{0.9cm} p{1.3cm} p{1.3cm} p{1.3cm}}
\toprule
    & & & &\multicolumn{3}{c}{Flower} & &\multicolumn{3}{c}{Pod} \\
\cmidrule(r){4-7} \cmidrule(r){8-11}
\multicolumn{2}{c}{Experiment} & Shift Domain & $\Delta$mAP (\%)& $mAP@50$ (\%) & Precision (\%) & Recall (\%) & $\Delta$mAP (\%)&
$mAP@50$ (\%) & Precision (\%) & Recall (\%) \\
\midrule\midrule
\multirow{1}{*}{%
  \parbox[c][0.6cm][c]{0.1cm}{%
    \centering\rotatebox{90}{$G_{s}E_{s}$}%
  }%
}
 & Seen L Y G & B & \deltacol{00.0} &76.3 & 80.8 & 71.5 &\deltacol{00.0} &65.1 & 76.2 & 59.2 \\
\specialrule{0.3pt}{15pt}{0pt}
\multirow{1}{*}{%
  \parbox[c][0.8cm][c]{0.1cm}{%
    \centering\rotatebox{90}{$G_{u}E_{s}$}%
  }%
}
 & Seen$\to$L Y, Unseen$\to$G & G &\deltacol{-1.7}& $\textbf{74.6}\pm2.08$ & $\textbf{78.6}\pm1.25$ & $\textbf{70.3}\pm2.80$ &\deltacol{-1.1}& $\textbf{64.0}\pm1.40$ & $\textbf{72.1}\pm0.80$ & $\textbf{60.3}\pm2.5$ \\
\specialrule{0.9pt}{15pt}{1pt}
\multirow{4}{*}{\rotatebox[origin=c]{90}{$G_{s}E_{u}$}}
 & Seen$\to$$Y_{2022}$ L G; Unseen$\to$$Y_{2023}$ & T &\deltacol{-10.6} &$\textbf{65.7}\pm4.74$ & $\textbf{73.9}\pm2.26$ & $\textbf{66.5}\pm3.25$ & \deltacol{-13.5}&$\textbf{51.6}\pm2.05$ & $\textbf{73.2}\pm2.62$ & $\textbf{55.0}\pm3.54$ \\
  & Seen$\to$$Y_{2023}$ L G; Unseen$\to$$Y_{2022}$ & T &\deltacol{-11.8}& $\textbf{64.5}\pm9.55$ & $\textbf{74.1}\pm0.42$ & $\textbf{66.2}\pm6.51$ & \deltacol{-14.9}&$\textbf{50.2}\pm1.41$ & $\textbf{79.4}\pm0.07$ & $\textbf{49.7}\pm2.40$ \\
 & Seen$\to$$L_{D}$ Y G; Unseen$\to$$L_{K}$  & S & \deltacol{-24.6} & $\textbf{51.7}\pm0.99$ & $\textbf{78.2}\pm1.91$ & $\textbf{58.1}\pm0.92$ &\deltacol{-13.8} &$\textbf{51.3}\pm1.48$ & $\textbf{78.7}\pm1.13$ & $\textbf{52.7}\pm5.23$ \\
  & Seen$\to$$L_{K}$ Y G; Unseen$\to$$L_{D}$  & S &\deltacol{-9.0}& $\textbf{67.3}\pm4.38$ & $\textbf{68.2}\pm1.13$ & $\textbf{72.8}\pm2.62$ &\deltacol{-16.9} &$\textbf{48.2}\pm0.00$ &  $\textbf{73.4}\pm0.14$ & $\textbf{49.9}\pm3.32$ \\
\specialrule{0.3pt}{1pt}{0pt}
\multirow{4}{*}{\rotatebox[origin=c]{90}{$G_{u}E_{u}$}}
 & Seen$\to$$L_{D}$ Y; Unseen$\to$$L_{K}$ G & SG &\deltacol{-25.7}& $\textbf{50.6}\pm3.82$ & $\textbf{75.2}\pm6.15$ & $\textbf{56.8}\pm5.09$ &\deltacol{-11.3} &$\textbf{53.8}\pm4.81$ & $\textbf{73.7}\pm4.03$ & $\textbf{57.0}\pm8.34$ \\
 & Seen$\to$$L_{K}$ Y; Unseen$\to$$L_{D}$ G & SG &\deltacol{-15.5}& $\textbf{60.8}\pm0.28$ & $\textbf{65.4}\pm1.84$ & $\textbf{69.5}\pm3.11$ & \deltacol{-20.1} &$\textbf{45.0}\pm2.76$ & $\textbf{66.6}\pm2.40$ & $\textbf{47.1}\pm3.96$ \\
 & Seen$\to$$Y_{2022}$ L; Unseen$\to$$Y_{2023}$ G & TG & \deltacol{-17.7} &$\textbf{58.6}\pm5.87$ & $\textbf{72.2}\pm4.53$ & $\textbf{60.4}\pm6.08$ & \deltacol{-16.7}&$\textbf{48.4}\pm1.20$ & $\textbf{75.4}\pm5.52$ & $\textbf{53.1}\pm2.40$ \\
 & Seen$\to$$Y_{2023}$ L; Unseen$\to$$Y_{2022}$ G& TG & \deltacol{-13.2} & $\textbf{63.1}\pm9.03$ & $\textbf{74.7}\pm2.55$ & $\textbf{65.6}\pm9.81$ & \deltacol{-19.2}&$\textbf{45.9}\pm0.78$ & $\textbf{72.6}\pm2.55$ & $\textbf{47.2}\pm1.27$ \\
\midrule\midrule
\end{tabular}
}
\footnotesize{L = Location, Y = Year, G = Genotype; B = Baseline; $L_{D}$ = Location at Davis, $L_{K}$ = Location at Kearney; T = Temporal, S = Spatial, SG = SpatioGenotype, TG = TemporalGenotype; Seen/Unseen indicate training exposure.}
\label{table:experiment results}
\end{table*}
For pod detection, the same patterns appear but with substantially larger sensitivity to distribution shifts. Under the baseline ($G_sE_s$), the model achieves 65.1\% mAP@50, 76.2\% precision, and 59.2\% recall, already lower than flower detection due to greater structural variability of pods. When evaluating unseen genotypes ($G_uE_s$), the drop is moderate (64.0\% mAP, $\Delta$mAP = -1.1 \%), again suggesting limited genotype-only effects. However, temporal shifts between years cause a large performance decrease, reducing mAP to 51.6–50.2\% with $\Delta$mAP values of $-13.5–14.9 \%$. Spatial shifts between Davis and Kearney show similarly strong degradation, producing 48.2–51.3\% mAP with $\Delta$mAP up to $-16.9 \%$, confirming substantial location-driven domain differences. When genotype variation is combined with environmental changes, performance declines even further. The spatio-genotype regime ($SG$) yields mAP values of 53.8\% and 45.0\%, while temporal-genotype shifts ($TG$) produce 48.4–45.9\% mAP and $\Delta$mAP up to $-20.1 \%$. While flower detection experiences a maximum degradation of $-25.7\%$ $\Delta$mAP, pod detection shows comparable or larger degradation under several regimes and consistently lower absolute accuracy. This difference likely reflects the greater visual and structural complexity of pods in field imagery. Compared with flowers, pods exhibit a wider range of appearances, are more frequently occluded by surrounding plant organs, and often resemble stems or other vegetative structures. Furthermore, pod images typically contain larger numbers of target objects, increasing overlap and scene complexity. Together, these factors may contribute to the lower baseline performance and greater sensitivity of pod detection to genotype-by-environment distribution shifts.

\subsection{Qualitative Analysis of Representation and Image Quality Shifts}
To further understand the mechanisms behind the performance degradation observed in Table~\ref{table:experiment results}, we analyze the underlying distributional differences between training and test datasets using both representation-level and image-level statistics. Specifically, we quantify shifts in the learned feature space using DINOv2 embeddings and evaluate variations in image quality characteristics that may contribute to domain shifts across genotype–environment conditions. This complementary analysis provides insight into why certain experimental regimes exhibit stronger generalization degradation than others.
\begin{figure*}[ht!]
    \centering
    \includegraphics[width=0.98\textwidth]{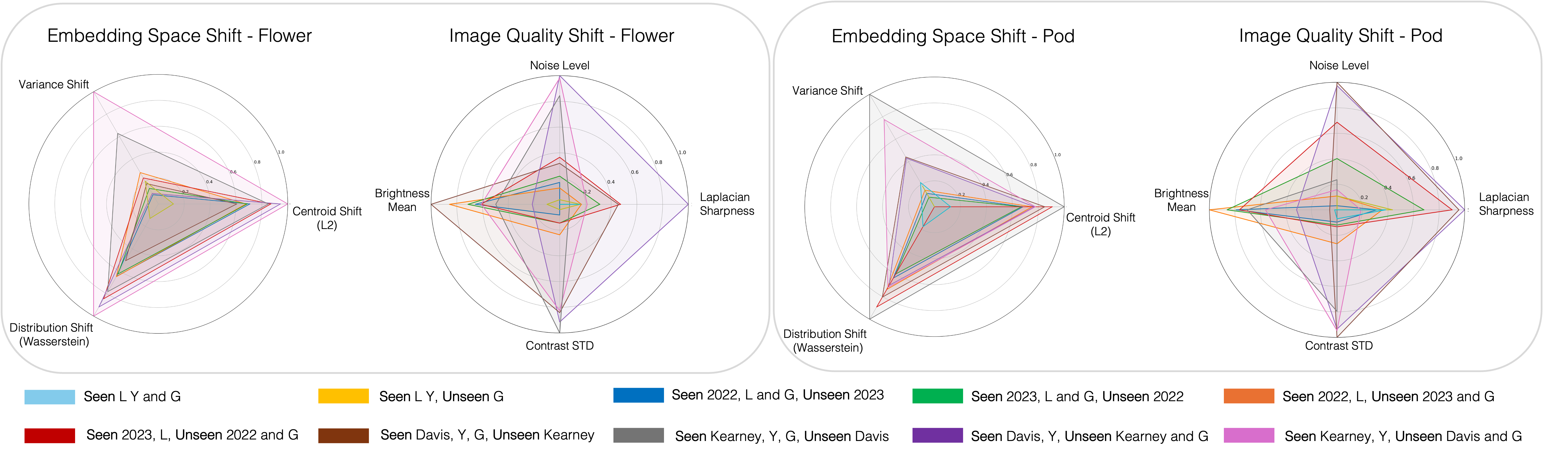}
    \caption{Quantification of domain shifts in the learned feature space using DINOv2 embeddings, alongside evaluation of variations in image quality characteristics. Radar plots summarize embedding-space shifts (e.g., centroid, variance, and distribution changes) and image quality metrics (e.g., noise, brightness, contrast, and sharpness) for both flower and pod datasets across experimental conditions.}
    \label{fig:shift_analysis}
\end{figure*}
\begin{table}[ht]
\centering
\footnotesize
\caption{Ranking of experiments based on total radar area (embedding + image quality shift). Higher area indicates larger domain shift severity.}
\resizebox{\textwidth}{!}{%
\begin{tabular}{l p{4.3cm} p{1.1cm} p{0.9cm} p{1cm} p{1.3cm} p{1.3cm} p{0.9cm} p{1cm} p{1.3cm} p{1.3cm}}
\toprule
    & & & &\multicolumn{3}{c}{Flower} & &\multicolumn{3}{c}{Pod} \\
\cmidrule(r){4-7} \cmidrule(r){8-11}
\multicolumn{2}{c}{Experiment} & Shift Domain & $\Delta$mAP (\%)& Rank & Emb. Area & ImgQ Area & $\Delta$mAP (\%)& Rank & Emb. Area & ImgQ Area \\
\midrule\midrule
\multirow{1}{*}{\parbox[c][0.6cm][c]{0.1cm}{\centering\rotatebox{90}{$G_{s}E_{s}$}}} & Seen L Y G                 & B & \deltacol{00.0} & 10 & 0.000 & 0.000 & \deltacol{00.0} & 10 & 0.000 & 0.024 \\ \specialrule{0.3pt}{15pt}{0pt}
\multirow{1}{*}{\parbox[c][0.8cm][c]{0.1cm}{\centering\rotatebox{90}{$G_{u}E_{s}$}}} & Seen$\to$L Y, Unseen$\to$G & G        &\deltacol{-1.7}  & 9  & 0.028 & 0.010 & \deltacol{-1.1} & 9  & 0.038 & 0.014 \\\specialrule{0.3pt}{15pt}{0pt}
\multirow{4}{*}{\rotatebox[origin=c]{90}{$G_{s}E_{u}$}}                              &Seen$\to$$Y_{2022}$ L G; Unseen$\to$$Y_{2023}$ & T &\deltacol{-10.6} & 8 & 0.246 & 0.083 & \deltacol{-13.5}& 8 & 0.256 & 0.071 \\
&Seen$\to$$Y_{2023}$ L G; Unseen$\to$$Y_{2022}$ & T &\deltacol{-11.8}& 7 & 0.265 & 0.187 & \deltacol{-14.9}& 6 & 0.220 & 0.402 \\
&Seen$\to$$L_{D}$ Y G; Unseen$\to$$L_{K}$ & S & \deltacol{-24.6} & 4 & 0.220 & 0.845 & \deltacol{-13.8} & 1 & 0.508 & 1.256 \\
&Seen$\to$$L_{K}$ Y G; Unseen$\to$$L_{D}$ & S &\deltacol{-9.0} & 3 & 0.733 & 0.530 & \deltacol{-16.9} & 4 & 0.745 & 0.413 \\  \specialrule{0.3pt}{1pt}{0pt}
\multirow{4}{*}{\rotatebox[origin=c]{90}{$G_{u}E_{u}$}}
 & Seen$\to$$L_{D}$ Y; Unseen$\to$$L_{K}$ G & SG &\deltacol{-25.7}& 2 & 0.451 & 1.161 & \deltacol{-11.3} & 3 & 0.613 & 1.012 \\
&Seen$\to$$L_{K}$ Y; Unseen$\to$$L_{D}$ G & SG &\deltacol{-15.5}& 1 & 1.299 & 0.744 & \deltacol{-20.1} & 2 & 1.299 & 0.363 \\
&Seen$\to$$Y_{2022}$ L; Unseen$\to$$Y_{2023}$ G & TG & \deltacol{-17.7} & 6 & 0.351 & 0.187 & \deltacol{-16.7}& 7 & 0.335 & 0.254 \\
&Seen$\to$$Y_{2023}$ L; Unseen$\to$$Y_{2022}$ G & TG & \deltacol{-13.2} & 5 & 0.490 & 0.278 & \deltacol{-19.2}& 5 & 0.350 & 0.683 \\
\bottomrule
\end{tabular}
}
\footnotesize{L = Location, Y = Year, G = Genotype; B = Baseline; $L_{D}$ = Location at Davis, $L_{K}$ = Location at Kearney; T = Temporal, S = Spatial, SG = SpatioGenotype, TG = TemporalGenotype; Seen/Unseen indicate training exposure.}
\label{tab:shift_ranking}
\end{table}
Table~\ref{tab:shift_ranking} summarizes the resulting ranking of experiments based on total radar area (embedding + image quality shifts) shown in Figure~\ref{fig:shift_analysis}, where larger areas indicate stronger distributional differences between training and test domains.

The qualitative shift analysis is consistent with the quantitative performance trends reported in Table~\ref{tab:shift_ranking}. Experiments with minimal distributional differences, such as the baseline ($G_sE_s$) and genotype-only shifts ($G_uE_s$), exhibit the smallest radar areas and correspondingly small detection degradation ($\Delta$mAP $< 2\%$). Temporal shifts between years show moderate increases in both embedding and image-quality divergence, which aligns with the observed performance drop of approximately $10-12\%$ in $\Delta$mAP for flower and $13-15\%$ in $\Delta$mAP for pod detection. In contrast, spatial shifts between Davis and Kearney produce significantly larger embedding distribution changes and strong image-quality variations particularly in brightness, contrast, and sharpness resulting in substantially larger radar areas and larger performance degradation ($9-25\%$ in $\Delta$mAP for flower and $13-17\%$ in $\Delta$mAP for pod detection). The largest shifts occur when genotype variation is combined with environmental changes ($SG$ and $TG$ regimes), where embedding divergence and image-quality differences are simultaneously amplified. These conditions correspond to the strongest domain shifts and the largest drops in detection accuracy observed in Table~\ref{table:experiment results}.

Comparing flower and pod detection further highlights the relationship between representation shift and task sensitivity. Pod detection experiments generally exhibit larger embedding and image-quality radar areas and lower detection performance across many of the evaluated generalization regimes. This pattern suggests that pod detection may be more sensitive to distribution shifts than flower detection within the environments examined in this study. One possible explanation is the greater visual complexity of pods in field imagery, including increased occlusion, similarity to other plant organs, and higher object density within images. However, these factors were not explicitly quantified in this work. Overall, the combined embedding and image-quality analyses indicate that distribution shifts in representation space and image appearance are associated with reduced model generalization across genotype–environment conditions.

\subsection{Synthetic Data and Few-Shot Adaptation}
\subsubsection{Evaluation of the Domain-Gap Optimization}
Before reporting downstream detection performance, we first evaluate the optimization procedure itself, i.e., how much the staged Wasserstein-distance search of Section~\ref{sec:aug} actually reduced the measured synthetic-to-real discrepancy, and what augmentation parameters it converged to for the JPG and EXR pipelines.
\begin{table}[ht]
\centering
\footnotesize
\caption{Selected domain-gap-optimized augmentation parameters (final round) for the JPG and EXR synthetic pipelines. Ranges denote the sampled interval applied per augmented image. A dash indicates the parameter was not part of that pipeline's search space.}
\begin{tabular}{lcc}
\toprule
Parameter & JPG (optimized) & EXR (optimized) \\
\midrule
Exposure / brightness scale   & [0.85, 1.22]  & [0.47, 0.91] \\
Contrast scale                & [0.70, 1.19]  & [0.65, 0.75] \\
Saturation scale               & [0.77, 1.84]  & [0.38, 1.49] \\
Gamma                          & [0.57, 1.02]  & -- (not needed, $p=0$) \\
White balance, R gain          & [0.85, 1.11]  & [0.86, 1.18] \\
White balance, G gain          & [0.98, 1.29]  & [0.92, 1.04] \\
White balance, B gain          & [0.71, 1.30]  & [0.71, 1.12] \\
Sensor noise $\sigma$ (DN)     & [0.99, 8.87]  & [3.01, 3.58] \\
Sharpen amount                 & [0.08, 0.66]  & [0.15, 0.29] \\
JPEG quality (output/simulated) & 84 (fixed)    & [66, 92] \\
\bottomrule
\end{tabular}
\label{tab:opt_params}
\end{table}
Table~\ref{tab:opt_params} summarizes the final optimized parameter ranges. Two patterns stand out. First, the optimizer converged on a wider saturation and noise range for JPG than for EXR, consistent with 8-bit imagery needing a larger photometric correction to compensate for information already lost to quantization. Second, the search assigned essentially zero weight to gamma adjustment for the EXR pipeline ($p_{\mathrm{gamma}}=0$), instead relying on the exposure and white-balance stages applied directly in linear radiance space. This is consistent with the HDR representation not requiring a separate nonlinear tone correction once exposure is handled physically, as described in Section~\ref{sec:aug}.
Across the staged search rounds analyzed here, the combined discrepancy loss (normalized Wasserstein distance over image-quality statistics and embedding-space distances) fell from $1.65$ to $1.22$ for the JPG pipeline (a $26\%$ reduction) and from $3.59$ to $2.72$ for the EXR pipeline (a $24\%$ reduction), with the largest single-stage gains coming from the JPEG-compression and output-simulation stage in both pipelines ($13$--$13.5\%$ of the total reduction each), followed by white-balance, noise, and sharpening. This indicates that compression and sensor-noise statistics, not tone or color alone, were the dominant contributors to the measured domain gap.

Notably, the final optimized loss remained higher in absolute terms for EXR ($2.72$) than for JPG ($1.22$), yet EXR-Optimized was the stronger or comparable synthetic representation for downstream few-shot detection in most conditions reported below (Section~\ref{sec:aug}, \textbf{H4}). This gap between the two evaluation criteria is expected rather than contradictory: the optimization objective is a distributional proxy over low-level image statistics and embedding similarity, not detection accuracy itself, and a representation can match the real domain less closely by this proxy while still preserving more of the radiometric detail a detector can exploit. We treat this as a necessary but not sufficient condition for transfer, and evaluate it directly against detection performance in the remainder of this section.

\subsubsection{Few-Shot Detection Performance Across Synthetic-Real Configurations}
\begin{figure*}[ht!]
    \centering
    \includegraphics[width=0.98\textwidth]{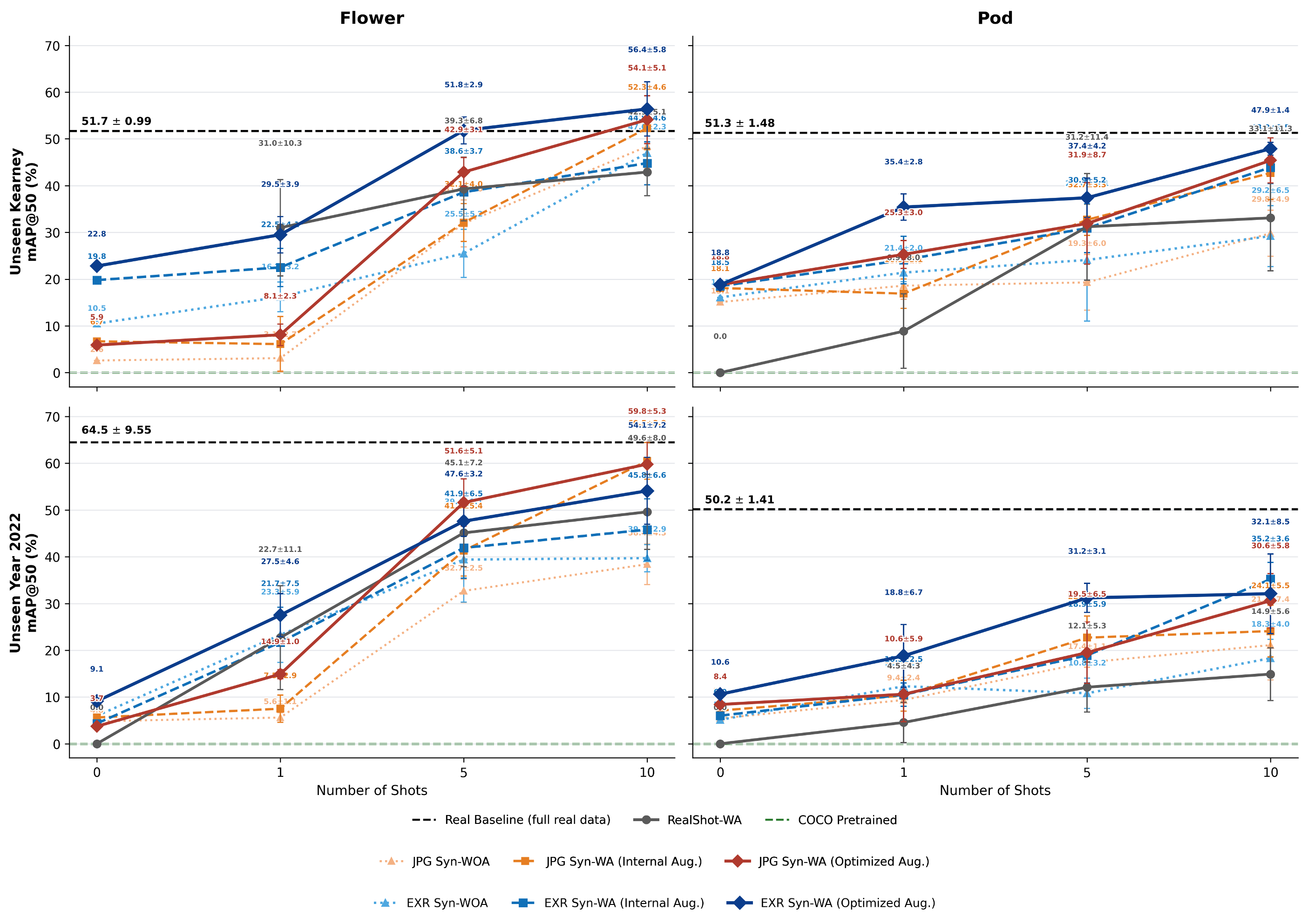}
    \caption{Detection performance under different scalability regimes using synthetic and real data for flower and pod detection. Results are reported as mAP@50 (mean $\pm$ std over 10 episodes) across varying few-shot settings (0, 1, 5, and 10 shots), with values labeled above each point. The left column presents flower detection results, while the right column shows pod detection results. The top row evaluates generalization to the unseen Kearney dataset, and the bottom row evaluates transferability to the unseen year 2022. Dashed black lines indicate the full real-data baseline, and dashed green lines represent the COCO-pretrained baseline. Grey curves correspond to real-only few-shot adaptation (\textit{RealShot-WA}). Warm-colored curves correspond to synthetic-real adaptation using the 8-bit JPG synthetic representation, and cool-colored curves correspond to the linear float32 HDR (EXR) representation, each shown without augmentation (\textit{Syn-WOA}, dotted), with hand-set internal augmentation (\textit{Syn-WA Internal}, dashed), and with domain-gap-optimized augmentation (\textit{Syn-WA Optimized}, solid).}
    \label{fig:syn_analysis}
\end{figure*}

Figure~\ref{fig:syn_analysis} summarizes detection performance under different few-shot and synthetic-data adaptation regimes for the two unseen target domains: unseen Kearney (spatial shift) and unseen year 2022 (temporal shift). For each domain and shot count ($k \in \{0,1,5,10\}$), we compare real-only few-shot supervision (\textit{RealShot-WA}) against six synthetic-plus-real configurations spanning two synthetic representations, 8-bit JPG and linear float32 HDR (EXR), and three augmentation states: no augmentation (\textit{Syn-WOA}), hand-set internal augmentation (\textit{Syn-WA Internal}), and domain-gap-optimized augmentation (\textit{Syn-WA Optimized}), as described in Section~\ref{sec:aug}. The full real-data baseline and the COCO-pretrained baseline are shown as reference lines in every panel. Complete numerical results for all environments and evaluation settings are provided in the supplementary materials (Table~S1). As in the $G \times E$ experiments above, the COCO-pretrained baseline produced essentially zero detection performance in every setting, confirming that generic object-detection pretraining alone is insufficient for this specialized task.

At zero real shots, where the synthetic-plus-real configurations reduce to synthetic-only transfer, augmentation quality and image representation both mattered substantially. Optimized augmentation consistently outperformed the corresponding unaugmented configuration, and the EXR representation was consistently the stronger of the two: zero-shot flower detection on unseen Kearney improved from $2.6$ mAP (JPG, no augmentation) to $22.8$ mAP (EXR, optimized augmentation), and zero-shot pod detection on unseen $Y_{2022}$ improved from $5.1$ mAP (EXR, no augmentation) to $10.6$ mAP (EXR, optimized augmentation). Across all four task-domain combinations, EXR-Optimized was the strongest or near-strongest zero-shot configuration, consistent with \textbf{H4}.

This advantage carried into the low-shot regime, particularly for pod detection. On unseen Kearney, one-shot pod detection reached only $8.87\pm7.97$ mAP with real-only supervision, compared with $35.4\pm2.81$ mAP for EXR-Optimized, a more than four-fold improvement with markedly lower variance across training episodes. A similar, though smaller, pattern was observed for flower detection: one-shot real-only supervision on unseen Kearney reached $31.0\pm10.3$ mAP, comparable in mean to EXR-Optimized ($29.5\pm3.9$ mAP) but with roughly three times the variance. This asymmetry, pod detection benefiting far more from synthetic supervision at low shot counts than flower detection, mirrors the greater intrinsic difficulty of pod detection identified in the $G \times E$ experiments (\textbf{H5}) and supports \textbf{H3}: synthetic supervision is most valuable precisely where real annotation is least sufficient on its own.

As the number of real shots increased to five and ten, the gap between augmentation states narrowed but did not close, and the best synthetic-real configurations matched or exceeded the full real-data baseline in several settings. On unseen Kearney, EXR-Optimized reached $51.8\pm2.9$ mAP at five shots and $56.4\pm5.8$ mAP at ten shots for flower detection, exceeding the real-data baseline of $51.7\pm0.99$ mAP at both points. JPG-Optimized reached $54.1\pm5.1$ mAP at ten shots, also above baseline. For pod detection on unseen Kearney, EXR-Optimized reached $47.9\pm1.36$ mAP at ten shots, within $3.4$ points of the $51.3\pm1.48$ baseline despite using only ten labeled real images. Gains were more modest for the temporal shift to unseen $Y_{2022}$: the strongest ten-shot flower configurations (JPG-Internal at $60.5\pm3.91$ mAP and JPG-Optimized at $59.8\pm5.35$ mAP) approached the $64.5\pm9.55$ baseline within its own standard deviation, while pod detection remained the most difficult combination overall, with the best ten-shot configuration (EXR-Internal, $35.2\pm3.56$ mAP) still substantially below the $50.2\pm1.41$ baseline. This last result indicates that the combination of temporal shift and pod detection is the regime least well addressed by the current synthetic pipeline, and a natural target for further gains in rendering realism or augmentation search.

Across nearly all conditions, optimized augmentation outperformed both the unaugmented and hand-set internal configurations, and EXR outperformed JPG, consistent with \textbf{H4}. This ranking was not perfectly monotonic at every shot count: JPG-Internal marginally exceeded JPG-Optimized at ten shots for flower detection on unseen $Y_{2022}$, and EXR-Internal exceeded EXR-Optimized at ten shots for pod detection on the same domain, indicating that the current Wasserstein-distance-based search improves the average image-quality and embedding-level match to the real domain but does not guarantee the optimal configuration at every operating point. Overall, these results support \textbf{H3}, demonstrating that synthetic data combined with few-shot real supervision substantially improves cross-domain detection performance and reduces the number of real annotations required to reach a given performance level, and support \textbf{H4}, showing that this benefit scales with how closely the augmented synthetic distribution is fit to the real domain, with the HDR (EXR) representation consistently providing additional headroom over 8-bit JPG.

\subsubsection{Sensitivity Analysis: Augmentation State, Format, and the Domain-Gap Objective}
\label{sec:sensitivity}
The results above raise a natural follow-up question: of the two design choices evaluated, augmentation state (WOA, Internal, Optimized) and synthetic representation (JPG, EXR), which actually drives the observed performance differences, and does that answer depend on shot count or task? We address this with a two-way ANOVA (Augmentation $\times$ Format) fit separately for each task and shot count, using the two real domains (unseen Kearney, unseen $Y_{2022}$) as replicates. Because this small-sample design (n=2 per cell) has limited power, we additionally report a Monte Carlo robustness check in which 10 pseudo-episodes per domain were drawn from each configuration's reported Normal($\mu$, $\sigma$) and re-analyzed (n=20 per cell). The two versions agree on every effect the small-sample test could resolve, and the Monte Carlo version additionally resolves effects the small-sample test lacked the power to detect. We report the Monte Carlo results below, with partial $\eta^2$ as the effect-size measure.
\begin{figure*}[ht!]
    \centering
    \includegraphics[width=0.95\textwidth]{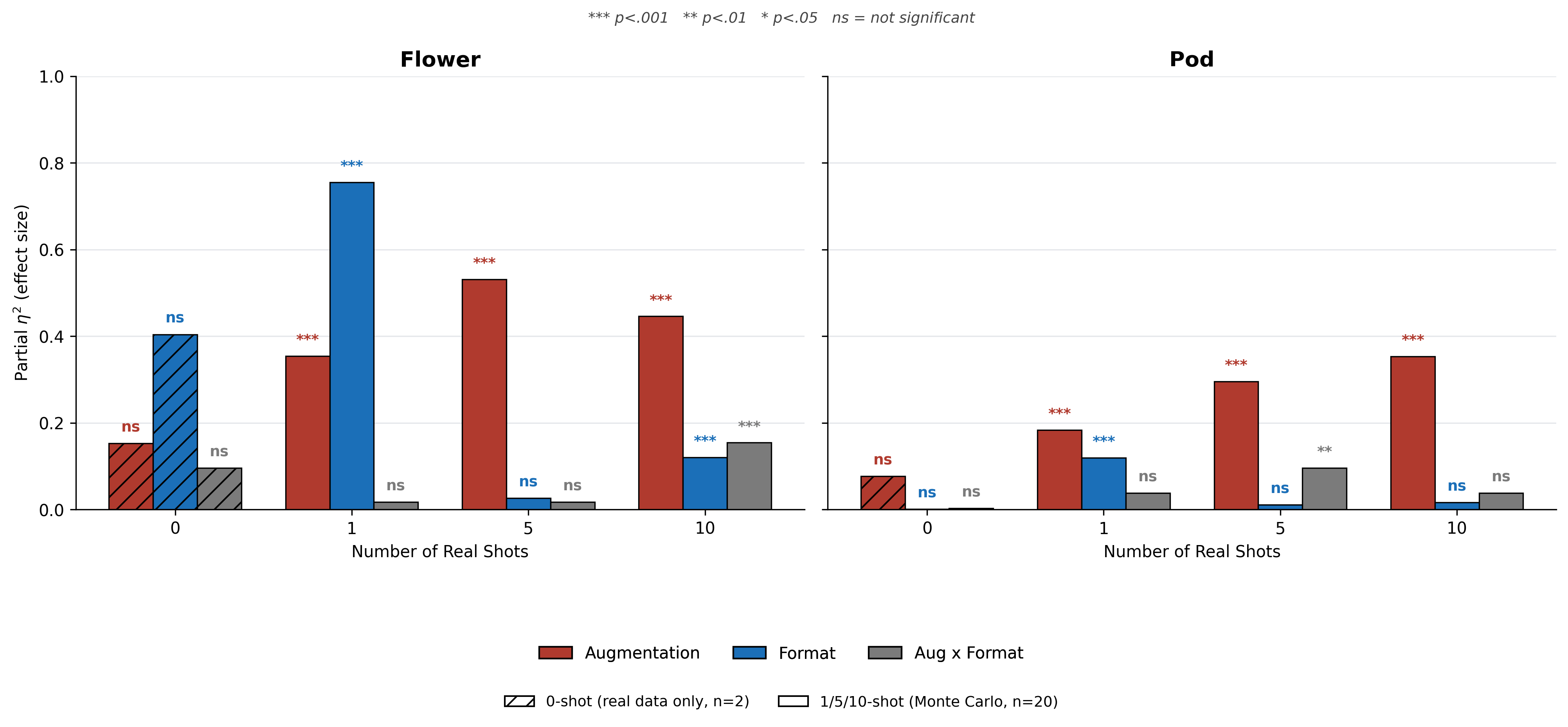}
    \caption{Sensitivity of flower and pod detection performance to Augmentation state, Format, and their interaction, expressed as partial $\eta^2$ from a two-way ANOVA fit separately at each shot count. Hatched bars (0-shot) use the real-data-only design (n=2 replicates, no reported variance to simulate from). Solid bars (1/5/10-shot) use the Monte Carlo-augmented design (n=20 replicates). Significance markers: *** p$<$.001, ** p$<$.01, * p$<$.05, ns = not significant.}
    \label{fig:anova_effects}
\end{figure*}
As shown in Figure~\ref{fig:anova_effects}, Augmentation state is a significant driver of detection performance for both tasks at every shot count once real images are available ($p<.001$ at 1, 5, and 10 shots, with $\eta^2_p$ ranging from .18--.53 for flower and .18--.35 for pod), but not at zero shots, where format dominates instead. Format shows the opposite pattern: it is a large, highly significant driver specifically at one shot ($\eta^2_p=.75$ for flower, $.12$ for pod, both $p<.001$), fades to non-significance by five shots for both tasks, and re-emerges as a smaller but significant effect for flower at ten shots ($\eta^2_p=.12$). The Augmentation $\times$ Format interaction is small and non-significant almost everywhere, with two exceptions, flower at ten shots and pod at five shots, that we treat as task-specific nuances rather than a general pattern. In short: \emph{which} synthetic representation you use matters most exactly where real supervision is scarcest, while \emph{how well the augmentation is tuned} matters continuously across the entire few-shot regime.
\begin{figure*}[ht!]
    \centering
    \includegraphics[width=0.95\textwidth]{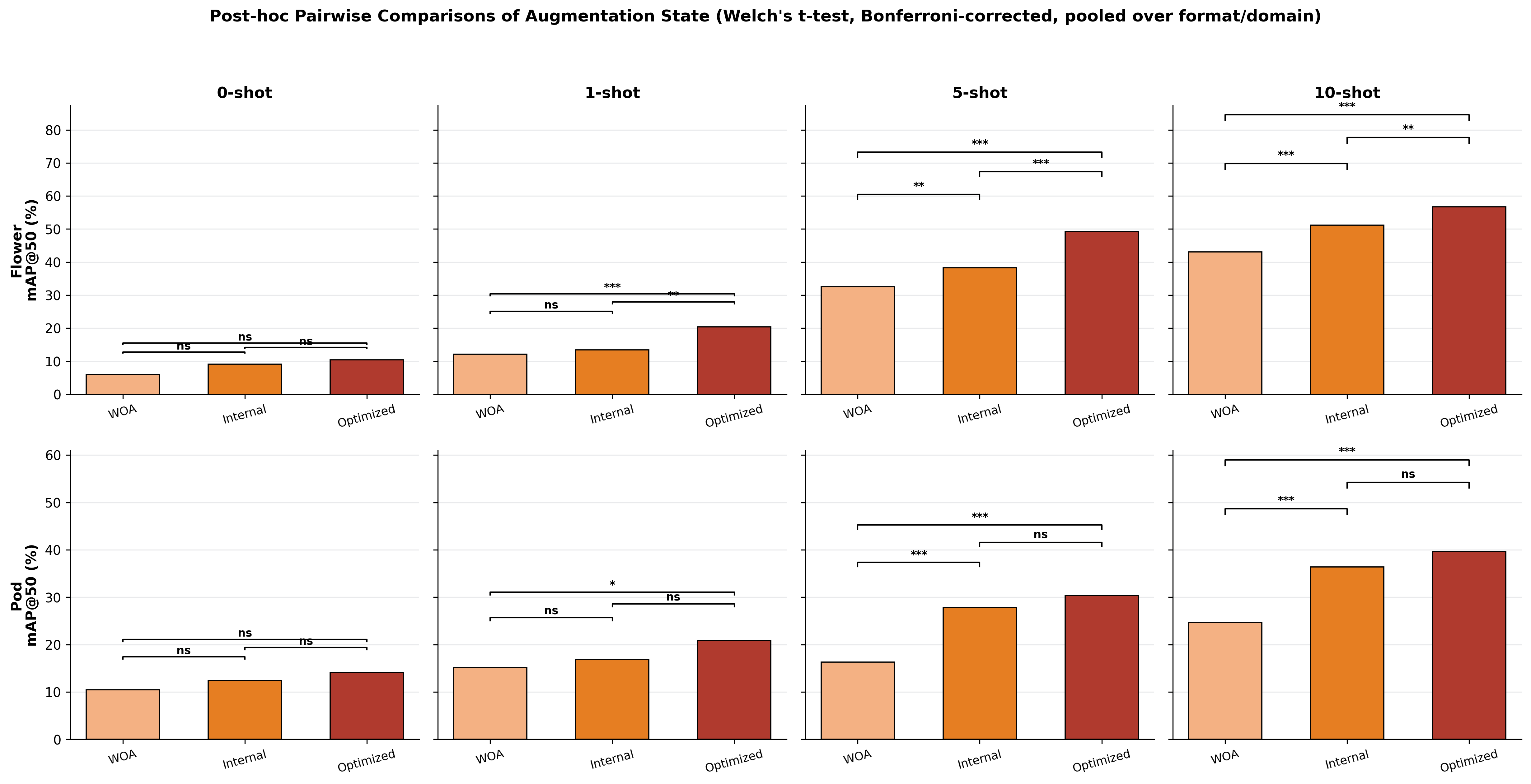}
    \caption{Post-hoc pairwise comparisons of Augmentation state (Welch's t-test, Bonferroni-corrected for three comparisons, pooled over format and domain) at each shot count. Significance brackets connect WOA vs. Internal, Internal vs. Optimized, and WOA vs. Optimized.}
    \label{fig:posthoc}
\end{figure*}

Because the omnibus ANOVA cannot say which specific augmentation states differ, we followed it with pairwise Welch's t-tests between the three augmentation states, pooled over format and domain and Bonferroni-corrected for three comparisons (Figure~\ref{fig:posthoc}). This reveals a task-dependent asymmetry that the aggregate result obscures. For flower detection, the Internal-to-Optimized step is itself a significant improvement at every shot count (1-shot: $p_{adj}=.0036$, 5-shot: $p_{adj}<.0001$, 10-shot: $p_{adj}=.0068$), indicating that the domain-gap optimization search adds real value beyond simply turning augmentation on. For pod detection, the same comparison is never significant at any shot count ($p_{adj}$ ranging .17--.61). Nearly all of pod's augmentation benefit is already captured by the WOA-to-Internal step, significant from 5 shots onward ($p_{adj}<.0001$), and the additional optimization step does not produce a statistically distinguishable further gain. Put directly, for flower, optimizing the augmentation is worth the extra engineering cost. For pod, hand-set augmentation already captures most of the achievable benefit, at least at the sample sizes evaluated here.
\begin{figure}[ht!]
    \centering
    \includegraphics[width=0.48\textwidth]{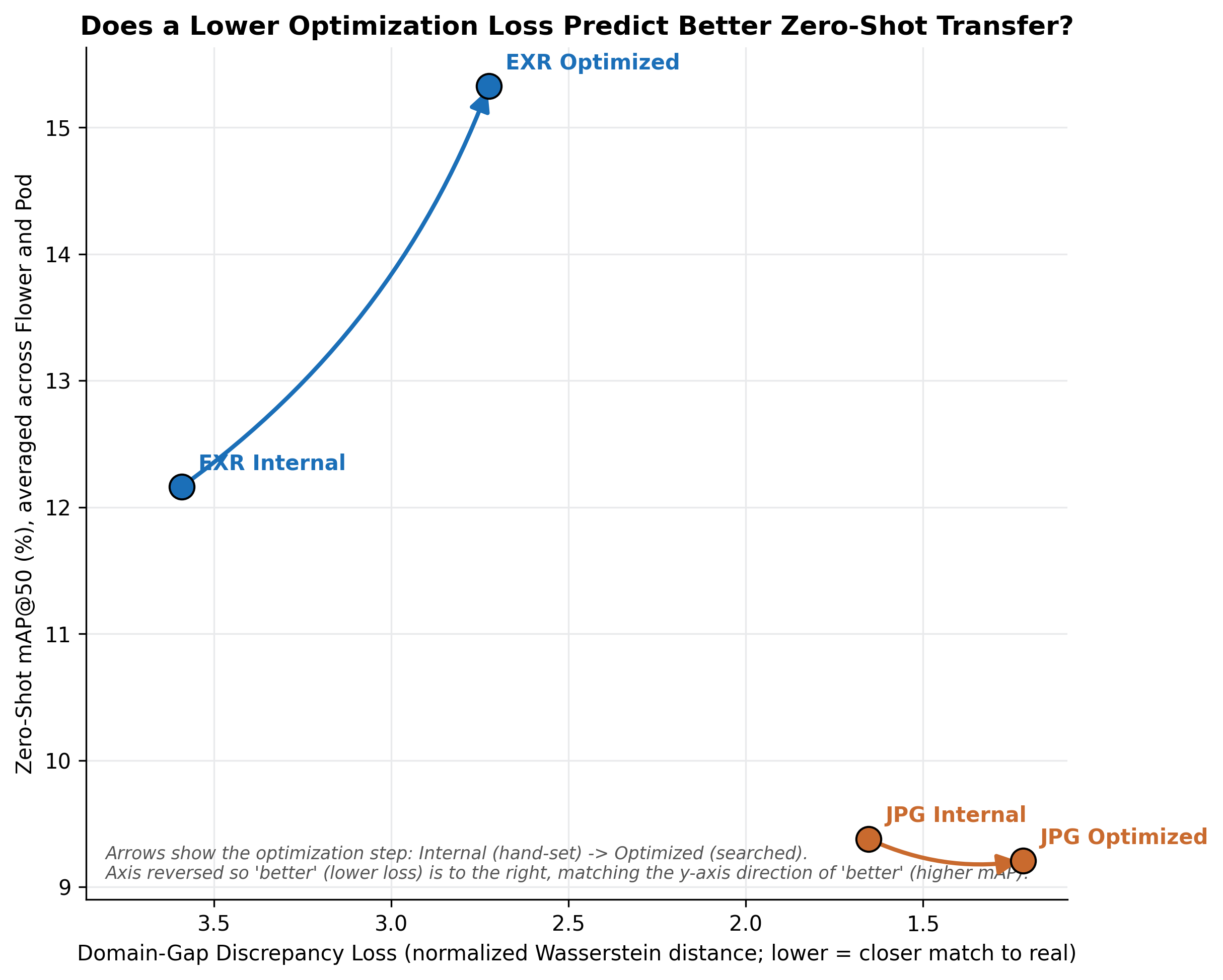}
    \caption{Relationship between the domain-gap optimization loss (Section~\ref{sec:aug}, lower is a closer statistical match to real imagery) and zero-shot detection performance (mAP@50, averaged across flower and pod), before and after optimization, for each synthetic representation. Arrows trace the optimization step (Internal $\to$ Optimized) within each format.}
    \label{fig:loss_vs_perf}
\end{figure}

This flower/pod asymmetry is clarified by comparing the two representations directly. Figure~\ref{fig:loss_vs_perf} plots the measured domain-gap discrepancy loss against zero-shot detection performance, before and after optimization, for JPG and EXR. For EXR, optimization reduced the loss from 3.59 to 2.72 (a 24\% reduction) and zero-shot mAP rose correspondingly, from 12.2 to 15.3 (a 26\% relative gain): the proxy objective and downstream performance moved together. For JPG, optimization reduced the loss by a comparable margin (1.65 to 1.22, 26\%), but zero-shot mAP did not follow (9.4 to 9.2, essentially flat). This is consistent with the radiometric argument made in Section~\ref{sec:aug}: because 8-bit JPG synthetic renders have already undergone an irreversible tone-mapping and quantization step, there is limited additional radiometric information left for a better-fit augmentation to exploit, whereas EXR's linear, unquantized representation gives the optimization search real headroom to translate a smaller measured domain gap into a detector-relevant improvement \cite{debevec1997recovering,brooks2019unprocessing}. We treat Figure~\ref{fig:loss_vs_perf} as illustrative of this mechanism rather than as a statistical test in its own right, given the small number of measured configurations.

Taken together, this sensitivity analysis refines \textbf{H4} rather than simply confirming it. Part (ii), that the HDR (EXR) representation yields a larger and more reliable improvement than 8-bit JPG, holds strongly and is concentrated exactly where \textbf{H3} predicts synthetic supervision should matter most: the one-shot, lowest-real-data regime. Part (i), that explicitly optimized augmentation outperforms hand-set augmentation, holds clearly for flower detection but not for pod detection, and the loss-versus-performance comparison suggests why: optimization can only convert a smaller measured domain gap into a detection gain when the underlying representation retains enough radiometric information to make that gap meaningful, which is more consistently true for EXR than for JPG. Practically, this indicates that the added engineering cost of domain-gap optimization (Section~\ref{sec:aug}) is best justified when paired with the HDR synthetic pipeline. For 8-bit synthetic imagery, and for object classes like pods where hand-set augmentation already captures most of the benefit, a simpler, non-optimized augmentation may be sufficient.

\subsection{Visual Analysis of Detection Successes and Failures}
\begin{figure*}[ht!]
    \centering
    \includegraphics[width=0.98\textwidth]{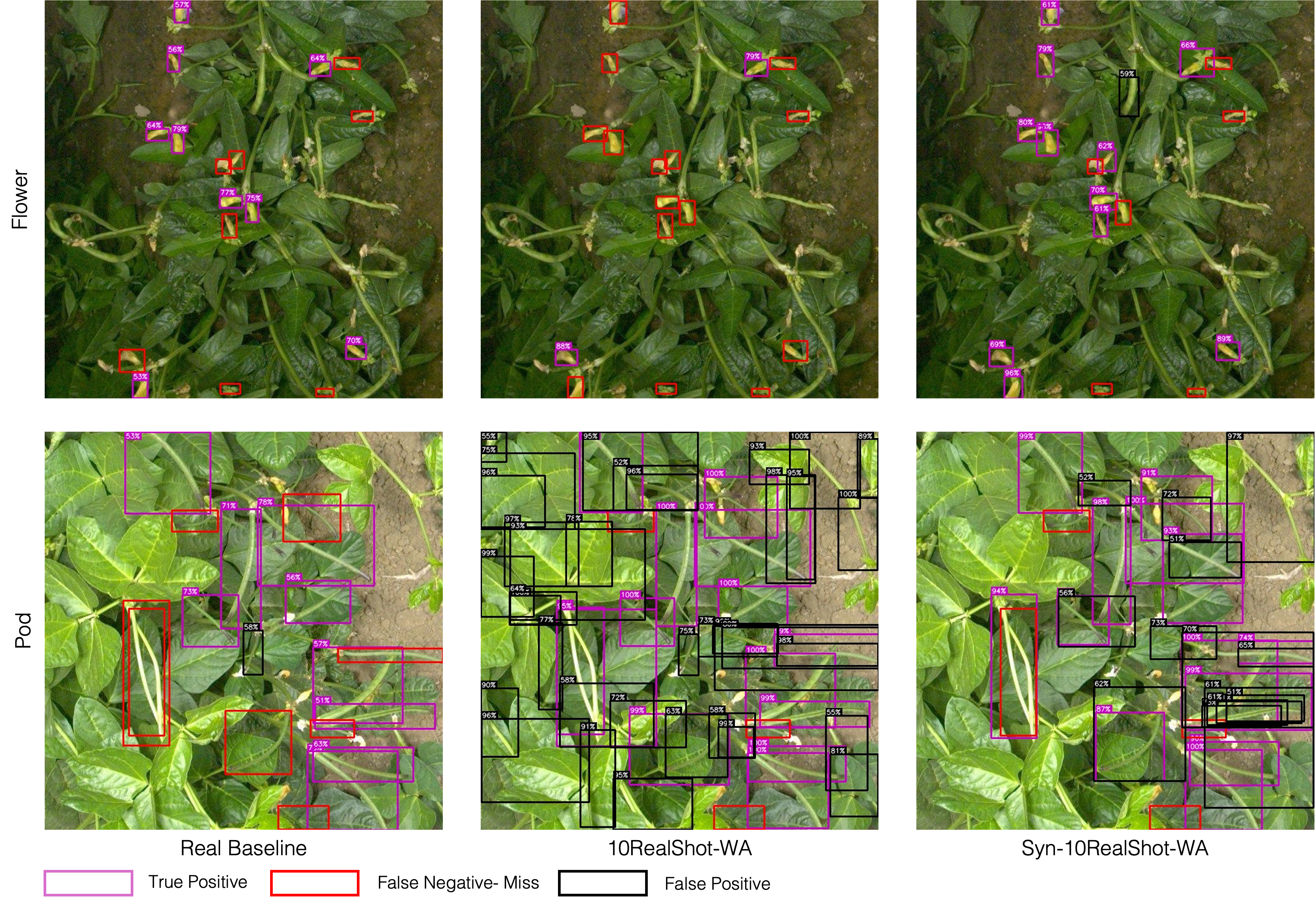}
    \caption{
    Qualitative comparison of prediction outputs under the unseen Kearney spatial generalization setting. The top row shows flower detection results, while the bottom row shows pod detection results. Columns correspond to the full real-data baseline, real-only few-shot adaptation (\textit{10RealShot-WA}), and synthetic-real adaptation (\textit{Syn-10RealShot-WA}). Purple boxes indicate true positives, red boxes indicate false negatives (missed detections), and black boxes indicate false positives.
    }
    \label{fig:prediction_analysis}
\end{figure*}

To further interpret these findings, Figure~\ref{fig:prediction_analysis} presents representative prediction outputs from the unseen Kearney ($L_K$) spatial generalization experiment under three training strategies: the full real-data baseline, real-only few-shot adaptation (\textit{10RealShot-WA}), and synthetic-real adaptation (\textit{Syn-10RealShot-WA}). The qualitative predictions closely mirror the quantitative trends observed in Figure~\ref{fig:syn_analysis} and provide visual evidence of how synthetic supervision improves robustness under domain shift.

The full real-data baseline produces the most stable predictions for both flower and pod detection, with relatively few false positives and strong localization accuracy. However, when the model is trained using only a limited number of real samples, prediction quality deteriorates substantially, particularly for pod detection. The \textit{10RealShot-WA} configuration exhibits a large number of false positives and unstable bounding-box localization, suggesting that limited real supervision alone is insufficient to capture the visual complexity of the unseen Kearney environment.
In contrast, incorporating synthetic supervision through the \textit{Syn-10RealShot-WA} strategy substantially improves prediction quality. For flower detection, the hybrid model produces more complete detections with fewer missed flowers and improved localization consistency compared with the real-only few-shot model. The improvement is even more apparent for pod detection, where synthetic-real adaptation significantly reduces excessive false positives while improving recall for elongated and partially occluded pods. These qualitative observations align with the quantitative improvements reported earlier, where \textit{Syn-10RealShot-WA} increased flower detection performance from $42.9$ to $55.6$ mAP and improved pod detection from $33.1$ to $36.1$ mAP under the unseen Kearney setting.

\section{Discussion}
The objective of this study was to quantify how genotype--environment ($G \times E$) distribution shifts affect the generalization of deep learning models for cowpea flower and pod detection, and to test whether synthetic data, properly aligned to the real domain, can overcome the resulting limits. By organizing the experiments into explicit $\mathbf{G_sE_s}$, $\mathbf{G_uE_s}$, $\mathbf{G_sE_u}$, and $\mathbf{G_uE_u}$ regimes, this work provides a structured framework for separating genotype effects, environmental effects, and their combined influence on high-throughput phenotyping model performance.
The results show that genotype-only shifts produced relatively small performance degradation for both flower and pod detection. In the $\mathbf{G_uE_s}$ regime, mAP declined only slightly compared with the seen genotype--seen environment baseline. This finding partially supports \textbf{H1}, indicating that unseen genotypes can affect detection accuracy, but genotype variation alone was not the dominant source of generalization loss in this dataset. Instead, the model remained relatively robust when environmental conditions were represented during training.
In contrast, environmental shifts caused substantially larger performance drops, strongly supporting \textbf{H2}. Both temporal shifts between years and spatial shifts between Davis and Kearney reduced detection performance, with location shifts generally producing the strongest degradation. These results suggest that changes in illumination, background, canopy structure, plant density, image quality, and field layout create visual domain shifts that are more difficult for the model to overcome than genotype variation alone. The asymmetric transfer between Davis and Kearney further indicates that some environments contain greater visual complexity or broader variability, affecting how well models trained in one site generalize to another.

The comparison between genotype and environmental generalization further clarifies the practical burden that environmental variation imposes. The relatively strong performance under unseen genotypes but seen environments suggests that training across diverse environments helps the model learn more transferable object representations, whereas genotype diversity alone provides a comparatively smaller generalization benefit. From a practical breeding perspective, this finding is important because it suggests that data collection should prioritize environmental diversity when annotation resources are limited: including more locations and years in the training set may improve generalization to new genotypes more effectively than increasing genotype diversity within a narrow environmental range. This observation also motivates the synthetic-data strategy evaluated later in this study, since procedurally generated synthetic imagery offers a scalable route to environmental diversity that does not require additional real-world data collection.

The shift analysis using DINOv2 embeddings and image quality metrics provides mechanistic support for the observed performance trends. Experiments with small embedding and image-quality radar areas, such as the baseline and genotype-only shift, showed limited mAP degradation. In contrast, spatial and combined genotype--environment shifts produced larger radar areas and stronger performance losses. This indicates that model failure under domain shift is not random, but is associated with measurable changes in representation space and low-level image characteristics such as brightness, contrast, sharpness, and noise. Therefore, embedding and image-quality diagnostics can serve as useful tools for identifying challenging deployment scenarios before full model retraining.

The results also confirm \textbf{H5}, showing that pod detection is more sensitive to $G \times E$ variation than flower detection. Across most regimes, pod detection achieved lower absolute accuracy and showed larger or comparable degradation under domain shift. This is likely because pods are more variable in shape, color, orientation, occlusion level, and developmental stage. Flowers are generally more visually distinct, while pods often blend with stems, leaves, and shadows. As a result, pod detection represents the more difficult phenotyping task and may require additional annotation, stronger augmentation, or object-specific model adaptation.

The synthetic and few-shot experiments directly support \textbf{H3} and \textbf{H4}. As in the $G \times E$ experiments, COCO-pretrained models alone failed to detect flowers and pods, confirming that generic object-detection pretraining is insufficient for this specialized agricultural task. Synthetic data alone (zero real shots) improved substantially over COCO pretraining, but the size of that improvement depended strongly on augmentation quality and image representation: optimized, domain-gap-aware augmentation consistently outperformed both unaugmented synthetic data and augmentation with hand-set internal parameters, and the linear HDR (EXR) representation consistently outperformed standard 8-bit JPG, directly supporting \textbf{H4}. The strongest results were obtained when optimized synthetic supervision was combined with a small number of real images: for spatial generalization to unseen Kearney, EXR-Optimized synthetic-real training matched or exceeded the full real-data baseline using as few as five real images, and pod detection, the more difficult of the two tasks, benefited disproportionately from synthetic supervision at very low shot counts, where real-only training was both weaker and far less stable across training episodes. These results support \textbf{H3}: combining synthetic supervision with limited real supervision outperforms either source alone and substantially reduces the number of real annotations required to reach a given performance level. Gains were more modest for temporal generalization to unseen $Y_{2022}$, particularly for pod detection, indicating that the current synthetic pipeline narrows but does not fully close the domain gap under every type of distribution shift.

The prediction examples further reinforce \textbf{H5} and \textbf{H3}. Pod detection remains inherently more difficult because pods exhibit stronger structural variability, overlap, occlusion, and background confusion, leading to higher rates of both false negatives and false positives under domain shift. At the same time, the synthetic-real adaptation strategy demonstrates that synthetic supervision improves feature representation learning and stabilizes detection behavior under limited-data conditions, increasing flower detection performance from $42.9$ to $55.6$ mAP and pod detection from $33.1$ to $36.1$ mAP under the unseen Kearney setting when ten real images were combined with synthetic supervision. These findings suggest that synthetic data is most effective not as a replacement for real imagery, but as a domain-gap-aware complement to it: a representation-initialization and low-data adaptation strategy that most benefits exactly the harder, more variable object class and the more severe distribution shifts identified earlier in this study.

Overall, this study demonstrates that robust agricultural object detection requires more than high baseline accuracy. A model that performs well under matched training and testing conditions may still fail under new environments. Beyond documenting this failure mode, this study also provides a concrete, domain-gap-aware strategy for mitigating it: procedurally generated synthetic imagery, combined with physically motivated camera-realism augmentation whose parameters are explicitly optimized against measured real-image statistics, substantially narrows the gap to full real-data performance using only a handful of labeled real images. $G \times E$-aware evaluation, representation- and image-quality-level shift diagnostics, and domain-gap-optimized synthetic data together provide a scalable pathway toward robust AI-based phenotyping across diverse breeding environments, and clarify the specific conditions, task difficulty, shift type, and augmentation fidelity, under which synthetic supervision can substitute for costly real-world annotation.

\section{Conclusion}
This study quantified the impact of genotype--environment ($G \times E$) distribution shifts on deep learning-based cowpea flower and pod detection, and tested whether synthetic data, properly aligned to the real domain, can overcome the resulting generalization limits. Using YOLOv11x across multiple locations, years, and genotypes, we showed that model performance is highest when training and testing conditions are aligned, but decreases when models are evaluated on unseen environments or combined unseen genotype--environment regimes.

The findings indicate that environmental variation is the dominant driver of performance degradation. Genotype-only shifts caused relatively small reductions in mAP, while temporal and especially spatial shifts produced much larger declines. This supports the conclusion that location- and year-specific imaging conditions, canopy structure, phenology, and image quality have a stronger effect on detection generalization than genotype variation alone.
Representation-level analysis using DINOv2 embeddings and image-quality metrics further showed that detection performance degradation corresponds to measurable distribution shifts between training and testing domains. Larger embedding and image-quality radar areas were generally associated with larger mAP reductions, confirming that domain shift can be quantified and interpreted before deployment.

The comparison between flower and pod detection showed that pods are more sensitive to genotype and environmental variation. Pod detection consistently remained more challenging because of greater structural variability, occlusion, and growth-stage dependence. This identifies pod detection as a key bottleneck for robust cowpea reproductive phenotyping.
Finally, synthetic data provided an effective, and in several settings sufficient, strategy for closing the performance gap created by genotype-by-environment shift. Synthetic data alone, without augmentation, improved over generic COCO pretraining but remained limited by a residual synthetic-to-real domain gap. Augmenting synthetic imagery using camera-realism transforms whose parameters were explicitly optimized against measured real-image statistics, rather than left unaugmented or hand-set, substantially narrowed this gap, and the linear HDR (EXR) synthetic representation consistently outperformed standard 8-bit JPG. Combining this optimized synthetic supervision with as few as five to ten labeled real images matched or exceeded the full real-data baseline for spatial generalization to unseen Kearney, and produced its largest relative gains for pod detection, the more difficult and lower-baseline of the two tasks, at the lowest shot counts. These results demonstrate that domain-gap-aware synthetic-real training can substantially reduce annotation requirements and support faster adaptation to new genotype--environment conditions, though gains were more modest for temporal generalization to unseen 2022, indicating that further improvements to rendering realism or augmentation search remain warranted for the most challenging shift-task combinations.

In summary, this work shows that AI-based phenotyping models should be evaluated and designed with explicit attention to $G \times E$ generalization, and that synthetic data, when paired with domain-gap-aware augmentation optimization rather than used as-is, provides a scalable pathway toward overcoming the resulting generalization limits. Environmental diversity, shift-aware diagnostics, physically motivated and empirically optimized synthetic-to-real alignment, and targeted few-shot adaptation together offer a practical route toward robust flower and pod detection across real-world breeding trials.
\printbibliography
\end{document}